\def\year{2021}\relax
\documentclass[letterpaper]{article} % DO NOT CHANGE THIS
\usepackage{aaai21}  % DO NOT CHANGE THIS
\usepackage{times}  % DO NOT CHANGE THIS
\usepackage{helvet} % DO NOT CHANGE THIS
\usepackage{courier}  % DO NOT CHANGE THIS
\usepackage[hyphens]{url}  % DO NOT CHANGE THIS
\usepackage{graphicx} % DO NOT CHANGE THIS
\usepackage{natbib}  % DO NOT CHANGE THIS AND DO NOT ADD ANY OPTIONS TO IT
\usepackage{caption} % DO NOT CHANGE THIS AND DO NOT ADD ANY OPTIONS TO IT
\usepackage{amsmath,amsfonts,amssymb}
\usepackage{graphicx}
\usepackage{textcomp}
\usepackage{gensymb}
\usepackage{enumitem}
\usepackage{comment}
\usepackage{multirow}
\usepackage{caption}
\usepackage{subcaption}
\usepackage[dvipsnames]{xcolor}
\usepackage[switch]{lineno}

\newcommand{\dataname}{\textsc{FixMyPose}}
\newcommand{\posecorrectiontask}{pose-correctional-captioning}
\newcommand{\poseretrievaltask}{target-pose-retrieval}

\newcommand\blfootnote[1]{%
  \begingroup
  \renewcommand\thefootnote{}\footnote{#1}%
  \addtocounter{footnote}{-1}%
  \endgroup
}

\newcommand{\vcenteredinclude}[1]{\begingroup
\setbox0=\hbox{\includegraphics[height=.028\textwidth]{#1}}%
\parbox{\wd0}{\box0}\endgroup}
\title{\dataname{}: Pose Correctional Captioning and Retrieval}
\author {
Hyounghun Kim\textsuperscript{*}, 
Abhay Zala\textsuperscript{*},
Graham Burri, 
Mohit Bansal\\}
\affiliations{
Department of Computer Science \\
University of North Carolina at Chapel Hill\\
\{hyounghk, aszala, ghburri, mbansal\}@cs.unc.edu
}
\begin{document}

\maketitle

\blfootnote{\textsuperscript{*}Equal contribution.}

\begin{abstract}
Interest in physical therapy and individual exercises such as yoga/dance has increased alongside the well-being trend, and people globally enjoy such exercises at home/office via video streaming platforms. However, such exercises are hard to follow without expert guidance. Even if experts can help, it is almost impossible to give personalized feedback to every trainee remotely. Thus, automated pose correction systems are required more than ever, and we introduce a new captioning dataset named \dataname{} to address this need. We collect natural language descriptions of correcting a ``current" pose to look like a ``target" pose. To support a multilingual setup, we collect descriptions in both English and Hindi. The collected descriptions have interesting linguistic properties such as egocentric relations to the environment objects, analogous references, etc., requiring an understanding of spatial relations and commonsense knowledge about postures. Further, to avoid ML biases, we maintain a balance across characters with diverse demographics, who perform a variety of movements in several interior environments (e.g., homes, offices). From our \dataname{} dataset, we introduce two tasks: the \posecorrectiontask{} task and its reverse, the \poseretrievaltask{} task. During the correctional-captioning task, models must generate the descriptions of how to move from the current to the target pose image, whereas in the retrieval task, models should select the correct target pose given the initial pose and the correctional description. We present strong cross-attention baseline models (uni/multimodal, RL, multilingual) and also show that our baselines are competitive with other models when evaluated on other image-difference datasets. We also propose new task-specific metrics (object-match, body-part-match, direction-match) and conduct human evaluation for more reliable evaluation, and we demonstrate a large human-model performance gap suggesting room for promising future work. Finally, to verify the sim-to-real transfer of our \dataname{} dataset, we collect a set of real images and show promising performance on these images.\footnote{Data/code publicly available: \url{https://fixmypose-unc.github.io}}

\end{abstract}

\begin{figure}[t]
    \centering
    \includegraphics[width=0.95\columnwidth]{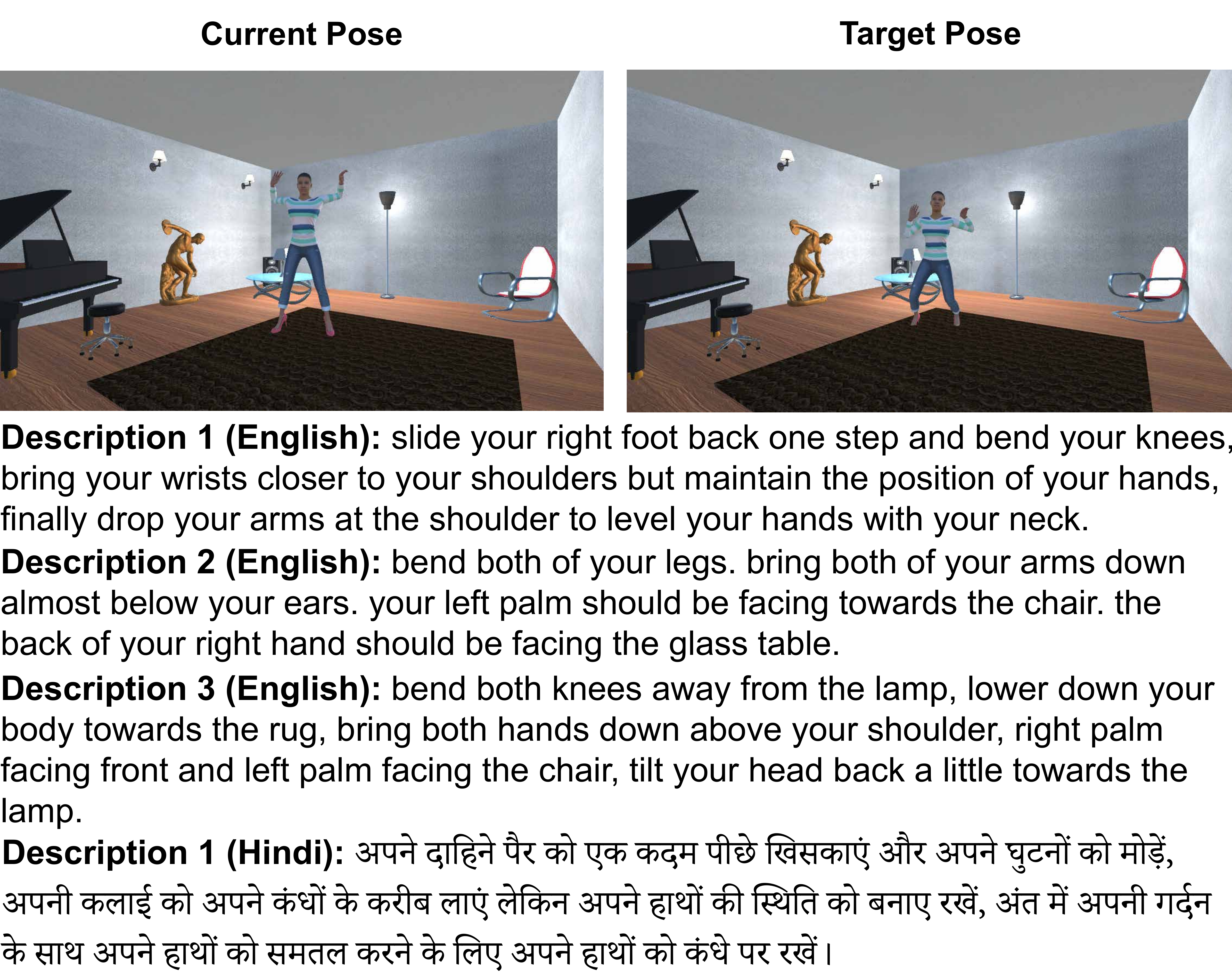}
    \vspace{-6pt}
    \caption{Current and target image pair and the corresponding correctional descriptions in both English and Hindi (we show only one of the three Hindi descriptions due to space).
    \label{fig:correct_capt_hindi}}
    \vspace{-6pt}
\end{figure}

\section{Introduction} 

As the well-being trend grows and people globally move to a new online lifestyle, interest in remotely (i.e., at home or in the office) learning health and exercise activities such as yoga, dance, and physical therapy is growing. Through advanced video streaming platforms, people can watch and follow the physical movements of experts, even without the expert being physically present (and hence scalable and less expensive). For such remote activities to be more effective, appropriate feedback systems are needed. For example, a feedback system should catch errors from the user's movements and give proper guidance to correct their poses. Related to this line of work, many efforts have been made on human pose estimation and action recognition~\cite{Johnson10,johnson2011learning,  andriluka14cvpr, Toshev_2014_CVPR, wei2016cpm, andriluka2018posetrack, yan2018spatial, zhao2019semantic, cao2019openpose, sun2019deep, verma2020yoga,  rong2020frankmocap}.
Research on describing the difference between multiple images has also been recently active~\cite{jhamtani2018learning, tan2019expressing, park2019robust, forbes2019neural}. 
However, there has been less focus on the human pose-difference captioning tasks, which require solving unique challenges such as understanding spatial relationships between multiple body parts and their movements. Moreover, the reverse task of retrieving or generating a target pose is also less studied. Combining these two directions together can allow for more interweaving human-machine communication in future automated exercise programs. 

Relatedly, interest in embodied systems for effective human-agent communication is increasing~\cite{kim2018does, wang2019exploring,abbasi2019multimodal, kim2020reducing}. Embodiment is also a desirable property when designing virtual assistants that provide feedback. For example, embodied virtual agents can show example movements to users or point at the users' body parts that need to move. Furthermore, for effective two-way communication with embodied agents, reverse information flow (i.e., human to agents) is also needed. A user may want to describe what actions they took so that the agent can confirm whether the user moved correctly or needs to change their movement. The agent should also be able to move its body to match the pose that the user is describing to help itself understand.

Therefore, to encourage the multimodal AI research community to explore these two tasks, we introduce a new dataset on detailed pose correctional descriptions called \dataname{} (\vcenteredinclude{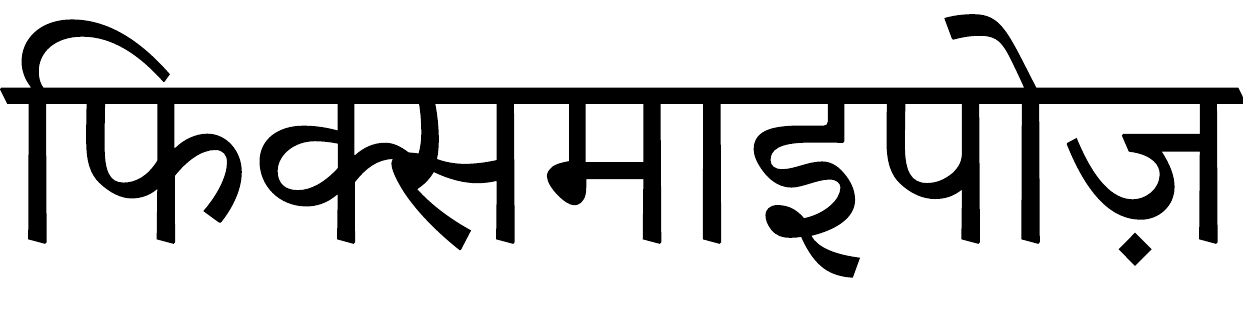}), which consists of image pairs (a ``current" and ``target" image) and corresponding correctional descriptions in both English and Hindi (Fig.~\ref{fig:correct_capt_hindi}). To understand our dataset, imagine you are in a physical therapy program following an instructor in a prerecorded video at home. Your movements and resulting pose are likely to be wrong, hence, you would like a virtual AI assistant to provide detailed verbal guidance on how you can adjust to match the pose of the instructor. In this case, your incorrect pose is in the ``current" image and the pose of the instructor is in the ``target" image, forming a pair. The verbal guidance from the virtual AI assistant is the correctional description.

From our \dataname{} dataset, we introduce two tasks for multimodal AI/NLP models: the `\posecorrectiontask{}' task and the `\poseretrievaltask{}' task. In the \posecorrectiontask{} task, models are given the ``current" and ``target" images and should generate a correctional description. The \poseretrievaltask{} task is the reverse of the \posecorrectiontask{} task, where models should select the correct ``target" image among other distractor images, given the ``current" image and description. This two-task setup will test AI capabilities for both important directions in pose correction (i.e., agents generating verbal guidance for human pose correction, and reversely predicting/generating poses given instructions), to enable two-way communication between humans and embodied agents in future research.
To generate image pairs, we implement realistic 3D interior environments (see Sec.~\ref{sec:dataset} for details). We also extract body joint data from characters to allow diverse tasks such as pose-generation (Fig.~\ref{fig:base_joint_configuration_and_dist}). We collect descriptions for these image pairs by asking annotators from a crowdsourcing platform to explain to the characters how to adjust their pose shown in the ``current" image to the one shown in the ``target" image in an instructional manner from the characters' egocentric view (see Table~\ref{tbl:property_frequnecies}). Furthermore, we ask them to refer to objects in the environment to create more detailed and accurate correctional descriptions, adding diversity and requiring models to understand the spatial relationships between body parts and environmental objects. The descriptions also often describe movement indirectly through implicit movement descriptions and analogous references (e.g., "like you are holding a cane") (see Sec.~\ref{ling_prop_section}), which means AI models performing this task should develop a commonsense understanding of these movements and references. To encourage multimodal AI systems to expand beyond English, we include Hindi descriptions as well (Fig.~\ref{fig:correct_capt_hindi}).

 \begin{figure}[t]
    \centering
    \includegraphics[width=0.92\columnwidth]{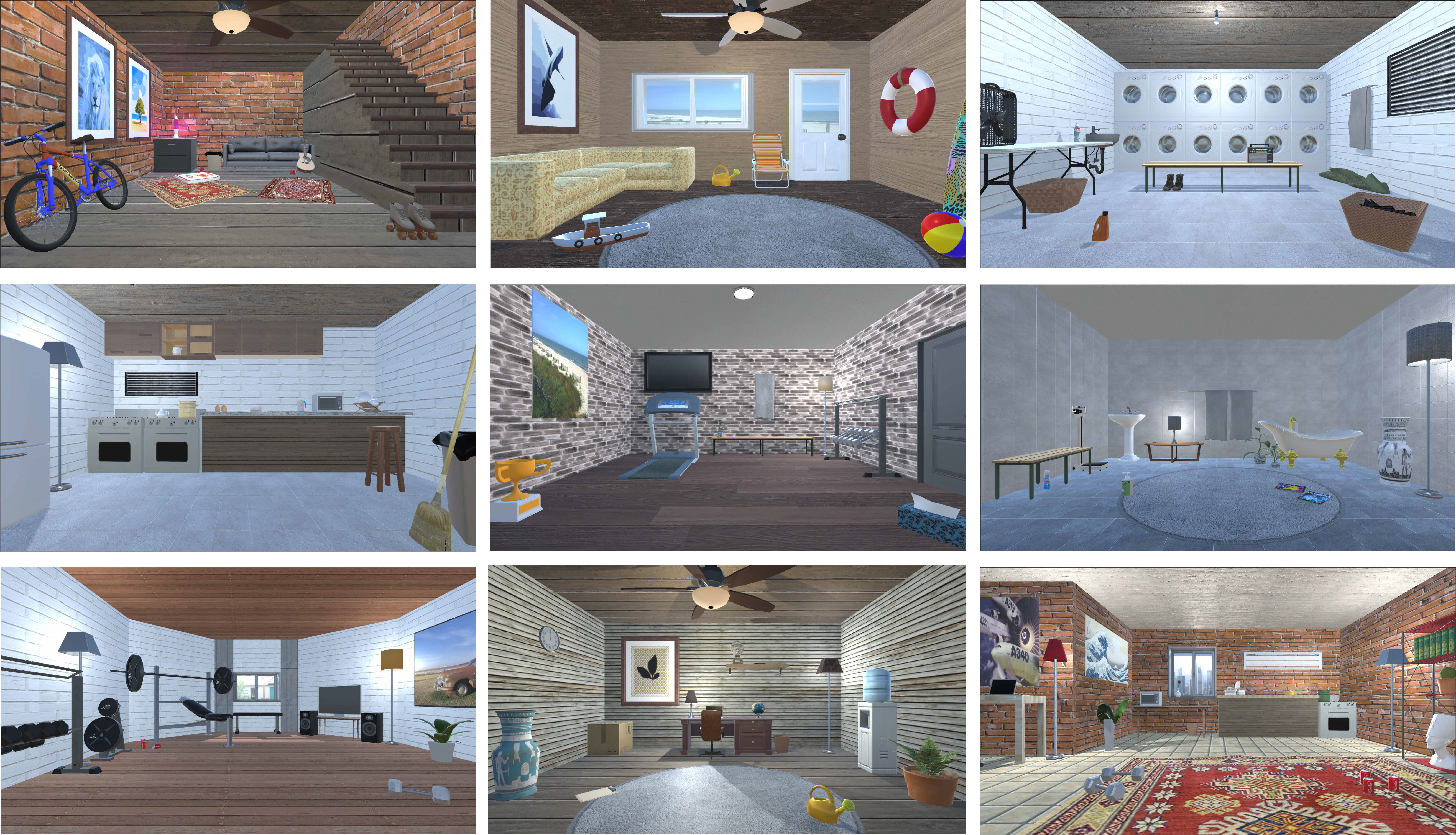}
    \caption{Example room environments: each room has a diverse style/theme (e.g., office, bathroom, living room).
    \label{fig:rooms}
    }
\end{figure}

Empirically, we present both unimodal and multimodal baseline models as strong starting points for each task, where we apply multiple cross-attention layers to integrate vision, body-joints, and language features. For the \posecorrectiontask{} model, we employ reinforcement learning (RL), which uses self-critical sequence training \cite{rennie2017self}, for further improvement. Also, we present the results from a multilingual training setup (English+Hindi) which uses fewer parameters by sharing model components, but shows comparable scores.

The multimodal models in both tasks show better performance than unimodal models, across both qualitative human evaluation and several of the evaluation metrics, including our new task-specific metrics: object, body-part, and direction match (details in Sec.~\ref{sec:fwd_task_results}). There is also a large human-model performance gap on the tasks, allowing useful future work on our challenging dataset. We also show balanced scores on demographic ablations, implying that our dataset is not biased toward a specific subset. Furthermore, our model performs competitively with existing works when evaluated on other image-difference datasets (Image Editing Request~\cite{tan2019expressing}, NLVR2~\cite{suhr2019corpus}, and CLEVR-Change~\cite{park2019robust}). Finally, to verify the simulator-to-real transfer of our \dataname{} dataset, we collect a test-real split which consists of real-world image pairs and corresponding descriptions, and show promising performance on the real images.

Our contributions are 3-fold: (1) We introduce a new dataset, \dataname{}, to encourage research on the integrated field of human pose, correctional feedback systems on feature differences with spatial relation understanding, and embodied multimodal virtual agents; (2) We collect a multilingual (English/Hindi) dataset; (3) We propose two tasks based on our \dataname{} dataset (\posecorrectiontask{} and \poseretrievaltask{}), and present several strong baselines as useful starting points for future work (and also demonstrate sim-to-real transfer).

\section{Related Work}
\noindent\textbf{Image Captioning.}
Describing image contents in natural language has been actively studied \cite{xu2015show, yang2016review, rennie2017self, lu2017knowing, anderson2018bottom, melas2018training, yao2018exploring}. This progress has been encouraged by the introduction of large-scale captioning datasets \cite{hodosh2013framing, lin2014microsoft, plummer2015flickr30k, krishna2017visual, johnson2016densecap, krause2017hierarchical}. Recently, more diverse image captioning tasks, which consider two images and describes the difference between them, have been introduced \cite{jhamtani2018learning, tan2019expressing, park2019robust, forbes2019neural}. However, to the best of our knowledge, there exists no captioning dataset about describing human pose differences. Describing pose difference or body movement requires detailed multi-focus over all body parts and understanding relations between them, introducing new challenges for AI agents. This kind of dataset is promising because of its potential real-world applications in activities such as yoga, dance, and physical therapy.

\vspace{3pt}
\par
\noindent\textbf{Human Pose.}
Human pose estimation and action recognition have been a long-standing topic in the research community \cite{Johnson10,johnson2011learning,  andriluka14cvpr, Toshev_2014_CVPR, wei2016cpm, andriluka2018posetrack, yan2018spatial, zhao2019semantic, cao2019openpose, sun2019deep, verma2020yoga,  rong2020frankmocap}. Recently, researchers are also focusing on generation tasks which generate a body pose sequence from an input of a different type from another modality such as audio or spoken language \cite{ shlizerman2018audio, tang2018dance, NIPS2019_8617, zhuang2020towards, saunders2020progressive}. 
However, there have been no research attempts on text generation based on pose correction. Thus, our novel \dataname{} dataset will encourage the community to explore this new direction. 

\vspace{3pt}
\par
\noindent\textbf{Spatial Relationships.}
Understanding spatial relationships between objects is an important capability for AI agents. Thus, the topic has attracted much attention from researchers~\cite{bisk2016towards, wang2016learning, li2016spatial, bisk2018learning}. Our \dataname{} dataset is rich in such reasoning about spatial relations with a variety of expressions (not only simple directions of left/right/up/down). Moreover, all the spatial relationships in the descriptions of the \dataname{} dataset are considered from the characters' egocentric perspective, requiring models to understand the characters' viewpoints.

\begin{figure}[t]
    \centering
    \includegraphics[width=0.95\columnwidth]{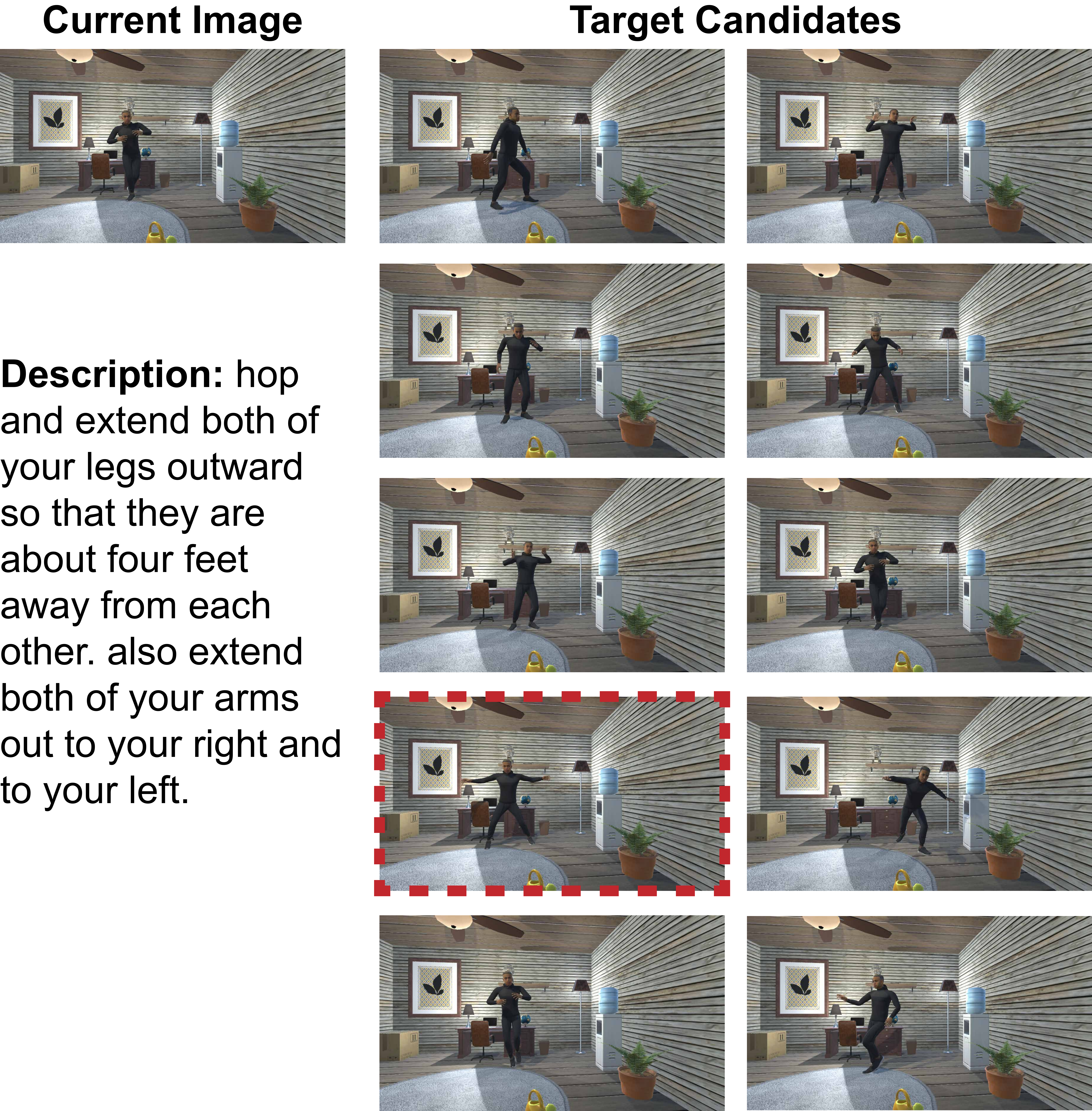}
    \caption{The \poseretrievaltask{} task: models have to select the correct ``target" image from a set of distractors (the image with red dashed border is the ground-truth target pose), given ``current" image and correctional description.}
    \label{fig:retrieval_task}
    
\end{figure}

\vspace{3pt}
\par
\noindent\textbf{Virtual Assistants.}
Virtual AI assistants such as Alexa, Google Assistant, Cortana, and Siri are already ubiquitous in our lives. However, there has been an increasing demand for multimodal (i.e., vision+language) virtual AI assistants, and as robotic and virtual/augmented/mixed reality technologies grow, so does interest in embodied virtual assistants~\cite{kim2018does, wang2019exploring, abbasi2019multimodal, kim2020reducing}. Our \dataname{} dataset will contribute to the evolution of embodied multimodal virtual assistants by providing a novel dataset as well as proposing a new approach on how to integrate physical movement guidance with virtual AI assistants.

\section{Tasks}

\vspace{3pt}
\par
\noindent\textbf{Pose Correctional-Captioning Task.}
During this task, the goal is to generate natural language (NL) correctional descriptions, considering the characters' egocentric view, that describe to a character how they should adjust their pose shown in the ``current" image to match the pose shown in the ``target" image (Fig.~\ref{fig:correct_capt_hindi}). As the ``current" and ``target" image pairs contain various objects in realistic room environments, models should have the ability to understand the spatial relationships between the body parts of characters and the environment from the characters' perspectives.
\vspace{3pt}
\par
\noindent\textbf{Target Pose Retrieval Task. \label{sec:pose_retri}}
Here, the goal is to select the correct ``target" image among 9 incorrect distractors, given the ``current" image and the corresponding correctional description (Fig.~\ref{fig:retrieval_task}). For the distractor images, we only consider images that are close to the ``target" pose in terms of body joints distances (see Appendix~\ref{app:distactor_choice} for detailed criteria). These distractor choices discourage models from easily discerning the correct ``target" image via shallow inference or shortcuts, requiring minute differences to be captured by models. The large human-model performance gap (Sec.~\ref{sec:bwd_task}) verifies the quality of our distractors.

\section{Dataset \label{sec:dataset}}

Our \dataname{} dataset is composed of image pairs with corresponding correctional descriptions in  English/Hindi.

\vspace{3pt}
\par
\noindent\textbf{Image, 3D Body Joints, and Environment Generation.}
We create 25 realistic 3D diverse room environments, filled with varying items (Fig.~\ref{fig:rooms}). To ensure diversity, we employ 6 human character avatars of different demographics across gender/race (each character is equally balanced in our dataset).\footnote{Our task focuses on understanding body movements/angles and not demographics, but we still ensure demographic diversity and balance in our dataset for ethical/fairness purposes so as to avoid unintended biases (e.g., see the balanced demographics ablation results and Sim-to-Real Transfer results on people with different demographics with respect to the 6 character avatars in Sec.~\ref{sec:results}). We plan to further expand our dataset with other types of diversity (e.g., height, age) based on digital avatar availability.} Since creating/modifying the body of characters requires 3D modeling/artistic expertise, we use pre-made character models that are publicly available (hence also copyright-free for our community's future use) in Adobe's Mixamo\footnote{https://www.mixamo.com}. In the rooms, the characters perform 20 movement animations and the camera captures images on a fixed interval. We also obtain 3D positional body joint data of the character's poses in the ``current" and ``target" images to provide additional useful features and allow a potential reverse pose-generation task (Fig.~\ref{fig:base_joint_configuration_and_dist}).
See Appendix~\ref{app:collection_details} for more on animation examples, environment creation, body joint data, and image capturing.

\begin{figure}
\centering
    \includegraphics[width=.9\columnwidth]{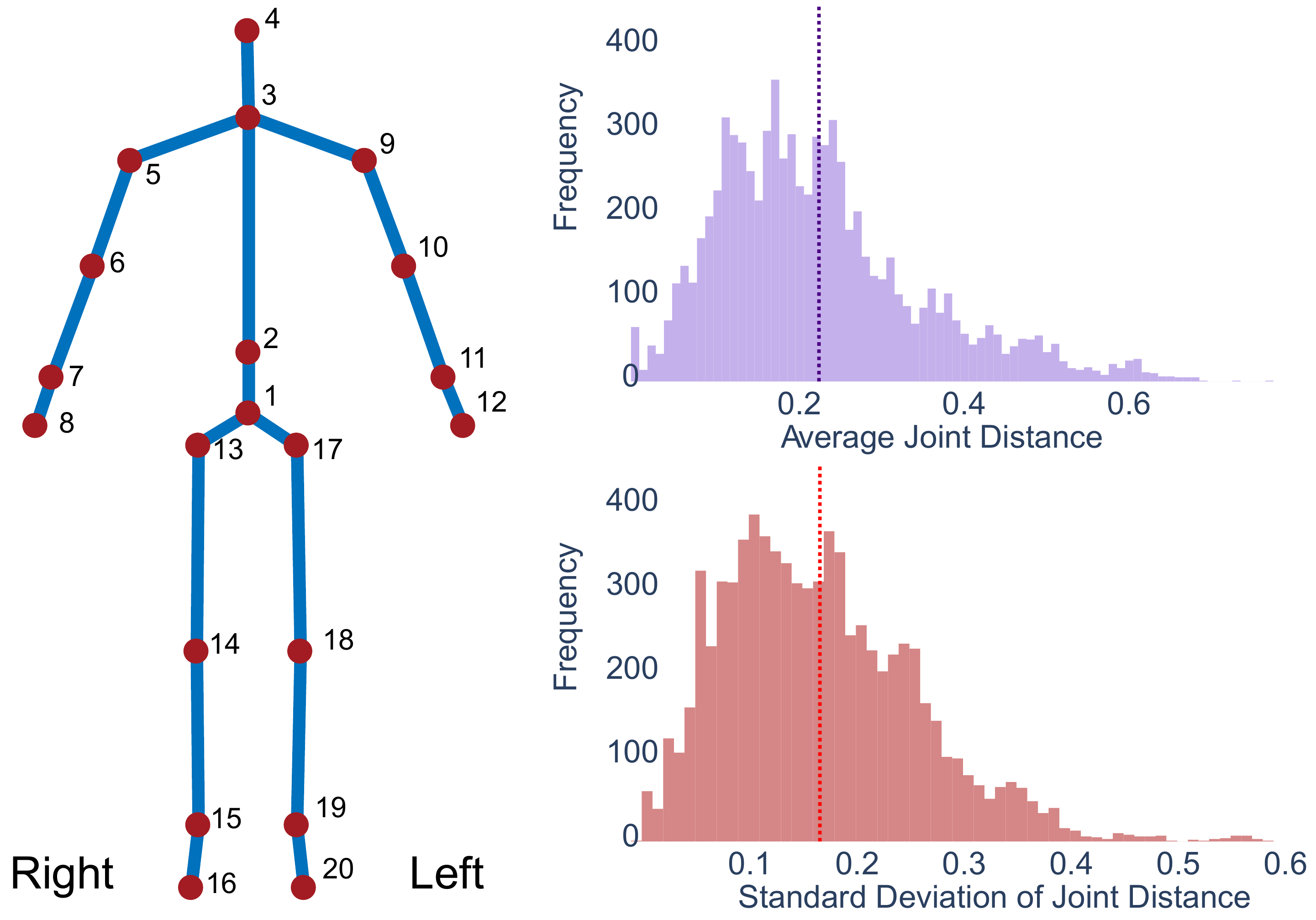}
    \caption{The 3D joint configuration of characters (left). The distribution of joint distances (meters) between poses of the ``current" and ``target" images (right). The Avg. of Min joint distances: 0.04 and the Avg. of Max joint distances: 0.65.}
    \label{fig:base_joint_configuration_and_dist}
\end{figure}

\vspace{3pt}
\par
\noindent\textbf{Description Collection.}
We employ annotators from the crowdsourcing platform Amazon Mechanical Turk\footnote{https://www.mturk.com} to collect the correctional descriptions. Workers are provided 3 images, ``current", ``target" images, and a ``difference" image that shows the difference between the two images, allowing them to write clear descriptions (see Appendix~\ref{app:dataset} for the images and collection interface). We ask them to write as if they are speaking to the characters as assistants who are helping them (like ``You should ...''), not calling them by the 3rd person (like ``The person ...'', ``They/She/He ...''). It also helps prevent accidental biased terms assuming the demographics of the characters. We collect 1 description for each image pair for the train split and 3 for all subsequent splits (i.e., val-seen/val-unseen/test-unseen) from unique workers, making the computation of automated evaluation metrics such as BLEU possible.

\vspace{3pt}
\par
\noindent\textbf{Description Verification.}
Each description and its corresponding image pair is given to a separate group of workers through a verification task. For each description, 3 different workers are asked to rank it from 1-4 based on its relevance to the image pair and its clarity, similar to previous works~\cite{lei2020tvr}. Descriptions that 2/3 of the workers rate lower than 3 are discarded. Image pairs that are flagged with certain issues are discarded as they do not provide good data (see Appendix~\ref{app:interfaces} for the verification interface and flags).

\vspace{3pt}
\par
\noindent\textbf{Hindi Data Collection.}
To collect the translated Hindi descriptions, we present a translation task to workers. Workers are given a description that has passed the verification task and its corresponding image pair to ensure the original meaning is not lost (see Appendix~\ref{app:interfaces} for the translation interface).

\vspace{3pt}
\par
\noindent\textbf{Worker Qualification and Payment.}
We require workers completing either of the tasks to be fluent in the needed languages and to have basic MTurk qualifications. The writing task takes around 1 minute and workers are paid \$0.18 per description. To encourage workers to write more and better descriptions, an additional increasing-bonus system is implemented. See Appendix~\ref{app:worker_qual_payment} for qualification/bonus/payment details.

\section{Data Analysis}
We collect 7,691 image pairs and 11,127 correctional descriptions for both English and Hindi (1 per train and 3 per evaluation splits). Our dataset size is comparable to other captioning tasks/datasets such as Image Editing Request~\cite{tan2019expressing} (3.9K image pairs/5.7K instructions), Spot-the-Diff~\cite{jhamtani2018learning} (13.2K image pairs/captions), and Birds-to-Words~\cite{forbes2019neural} (3.3K image pairs/16K paragraphs). We plan to keep extending the dataset and add other languages in the future.

\subsection{Statistics}
\noindent\textbf{Joint Distances.}
Fig.~\ref{fig:base_joint_configuration_and_dist} shows the distribution of average joint distances (meters) between the poses in the ``current" and ``target" images. As indicated by the mean (0.24m), stddev (0.18m), and min/max (0.04/0.65m) of the average distance of individual joints, models should be able to capture different movement levels simultaneously in an image pair.

\vspace{3pt}
\par
\noindent\textbf{Description Vocabulary and Length.}
The collection of descriptions in our \dataname{} dataset has 4,045/4,674 unique English/Hindi words. The most common words in both languages (see Appendix~\ref{app:most_common_pie} for details and pie charts) relate to direction, body parts, and movement, showing that models need to have a sense of direction with respect to body parts and objects, and also capture the differences between the poses to infer the proper movements. The average length of the multi-sentenced descriptions (49.25/52.74 words) is high, indicating that they are well detailed (see Appendix~\ref{app:description_length} for details).

\vspace{3pt}
\par
\subsection{Linguistic Properties}
\label{ling_prop_section}
To investigate the diverse linguistic properties in our dataset, we randomly sample 50 descriptions and manually count occurrences of traits. We found interesting traits (see Table~\ref{tbl:property_frequnecies} and Appendix~\ref{app:linguistic_properties} for examples), requiring agents to deeply understand characters' movements and express them in an applicable form (the Hindi descriptions also share these traits).

\begin{table}[t]
\begin{center}
\resizebox{0.98\columnwidth}{!}{
 \begin{tabular}{|c|c|c|c|}
 \hline
    Reference Frame & Freq. & Example (English)\\
 \hline
   \multirow{2}{*}{\shortstack[c]{Egocentric\\Relation}} & \multirow{2}{*}{100\%} & \multirow{2}{*}{\shortstack[c]{``... \textbf{rotate your left shoulder} so that\\ your hand is \textbf{above your elbow} ..."}} \\
    & & \\
    \hline
   \multirow{2}{*}{\shortstack[c]{Environmental\\Direction}} &  \multirow{2}{*}{52\%} & \multirow{2}{*}{\shortstack[c]{``... turn your left leg and right leg to the \\left to \textbf{face the wall with the door} ..."}} \\
    & & \\
   \hline
   \multirow{2}{*}{\shortstack[c]{Implicit Movement\\Description}} &  \multirow{2}{*}{58\%} & \multirow{2}{*}{\shortstack[c]{``... lean your body towards and\\ slightly over your right leg ..."}} \\
    & & \\
    \hline
    \multirow{2}{*}{\shortstack[c]{Analogous\\Reference}} &  \multirow{2}{*}{18\%} & \multirow{2}{*}{\shortstack[c]{``... in front of you \textbf{as if you are} \\\textbf{gesturing for someone to stop} ..."}} \\
     & & \\
   \hline
\end{tabular}
}
\end{center}
\caption{Examples of linguistic properties in correctional descriptions (see Appendix~\ref{app:linguistic_properties} for examples and image examples of implicit movement description). \label{tbl:property_frequnecies}
} 
\end{table}

\vspace{3pt}
\par
\noindent\textbf{Egocentric and Environmental Direction.}
Descriptions in our \dataname{} dataset are written considering the egocentric (first-person) view of the character. Descriptions also reference many environmental objects and their relation to the characters' body parts, again from an egocentric view. This means models must understand spatial relations of body parts and environmental features from the egocentric view of the character rather than the view of the ``camera".

\vspace{3pt}
\par
\noindent\textbf{Implicit Movement Description and Analogous Reference.}
Implicit movement description and analogous reference are often present in descriptions. These descriptions imply movements without needing to say them. Analogous references are a more extreme form of implicit movement description, where the movement is wrapped in an analogy. Models must develop commonsense knowledge of these movements in order to understand their meaning. See Table~\ref{tbl:property_frequnecies} and Appendix~\ref{app:linguistic_properties} for examples.

\section{Models}
\begin{figure}[t]
    \centering
    \includegraphics[width=0.87\columnwidth]{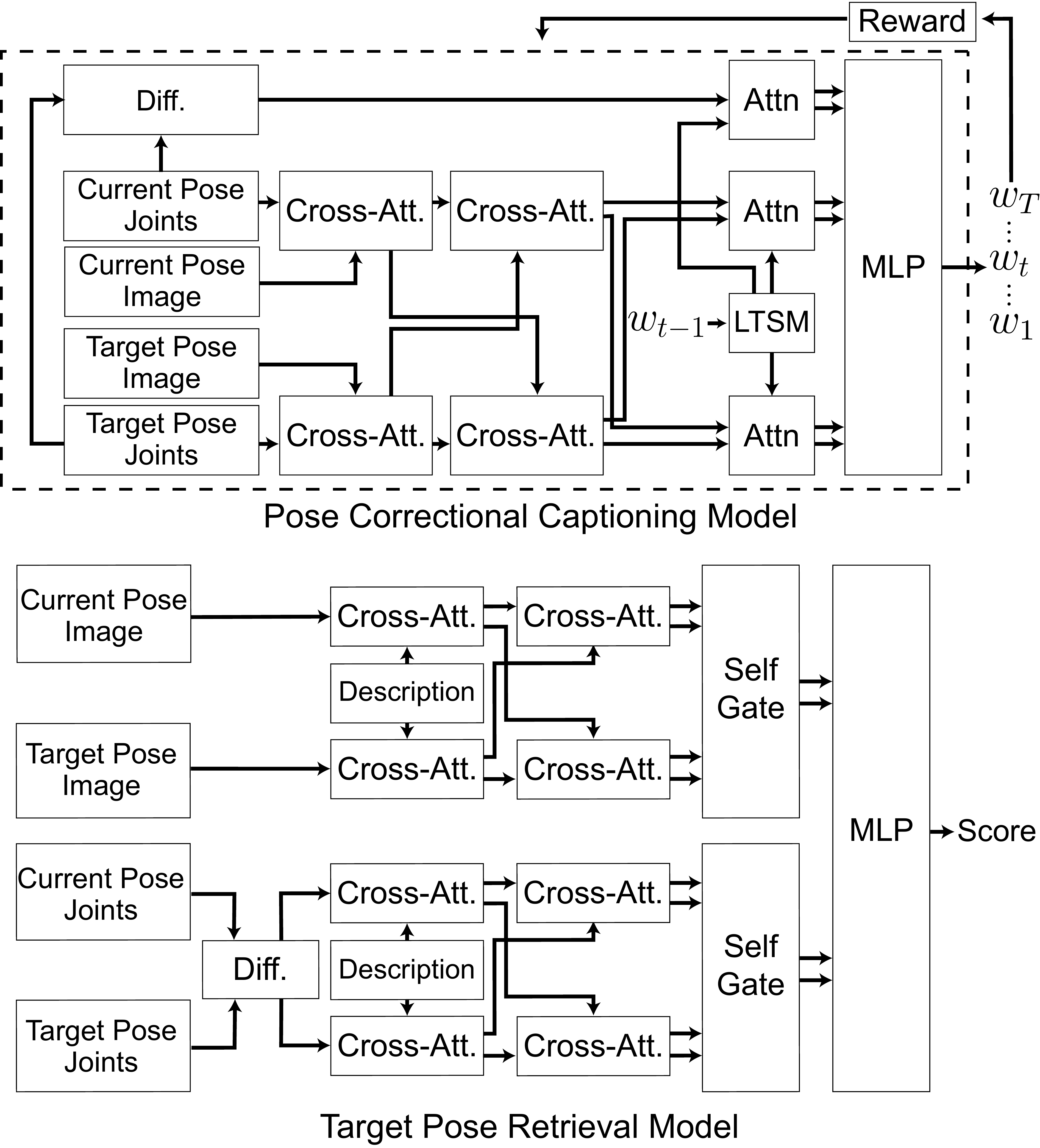}
    \caption{The \posecorrectiontask{} model (top) and the \poseretrievaltask{} model (bottom).\label{fig:model}}
\end{figure}

We present multiple strong baselines for both the \posecorrectiontask{} and \poseretrievaltask{} task (Fig.~\ref{fig:model}) to serve as starting points for future work.

\subsection{Pose Correctional Captioning Model}
We employ an encoder-decoder model for the \posecorrectiontask{} task. Also, we apply reinforcement learning (RL) after training the encoder-decoder model, and present multilingual training setup which reduces the number of parameters through parameter sharing.
\vspace{3pt}
\par
\noindent\textbf{Encoder.}
We employ ResNet \cite{he2016deep} to obtain visual features from images. To be specific, we extract feature maps \(f^c\) and \(f^t \in \mathbb{R}^{N\times N \times 2048} \) from the ``current" pose image \(I^c\) and the ``target" pose image \(I^t\), respectively: $ f^c = \textrm{ResNet}(I^c)$; $f^t = \textrm{ResNet}(I^t)$.
For 3D joints, \(J^{c}, J^{t} \in \mathbb{R}^{20\times3} \), we use linear layer to encode: $\hat{J^c} = \textrm{PE}(W_j^{\top}J^c)$; $\hat{J^t} = \textrm{PE}(W_j^{\top}J^t)$; $J^d = \textrm{PE}(W_j^{\top}(J^t-J^c))$, where \(W_j\) is the trainable parameter (all \(W_*\) from this point on denote trainable parameters) and \(\textrm{PE}\) \cite{gehring2017convolutional, vaswani2017attention} denotes positional encoding.
\vspace{3pt}
\par
\noindent\textbf{Decoder.}
Words from a description, \(\{w_t\}_{t=1}^T\), are embedded in the embedding layer: $\hat{w}_{t-1} = \textrm{Embed}(w_{t-1})$, then sequentially fed to the LSTM layer~\cite{hochreiter1997long}: $h_{t} = \textrm{LSTM}(\hat{w}_{t-1}, h_{t-1})$. We employ the bidirectional attention mechanism \cite{seo2016bidirectional} to align image features and joints features.
\begin{align}
     \tilde{f}^c, \tilde{J}^t, \tilde{f}^t, \tilde{J}^c &= \textrm{CA-Stack}(f^c, \hat{J}^c, f^t, \hat{J}^t)
\end{align}
where \(\textrm{CA-Stack}\) is a cross attention stack (see Appendix~\ref{app:model}).
\begin{align}
    f = W_c^{\top}[&\tilde{f}^c;\tilde{f}^t; \tilde{f}^c \odot \tilde{f}^t],\;
    J = W_c^{\top}[ \tilde{J}^c;\tilde{J}^t; \tilde{J}^c \odot \tilde{J}^t]\\
    f_t = \textrm{Att}(&h_t, f),\; J_t = \textrm{Att}(h_t, J),\; J_t^d = \textrm{Att}(h_t, J^d) \\
    k_t &= W_k^{\top}[f_t;J_t;h_t;h_t\odot f_t;h_t\odot J_t] \\
    g_t &= W_s^{\top}[k_t;J_t^d]
\end{align}
where \(\textrm{Att}\) is general attention (see Appendix~\ref{app:model} for details). The next token is: $w_t = \textrm{argmax}(g_t)$, and the loss is: $L_{ML} = -\sum_t\log{p(w_t^*|w_{1:t-1}^*, f, J)}$, where $w_t^*$ is the GT token.
\vspace{3pt}
\par
\noindent\textbf{RL Training.} We apply the REINFORCE algorithm~\cite{williams1992simple} to learn a policy $p_{\theta}$ upon the model pre-trained with the maximum likelihood approach: $L_{RL} = -\mathbb{E}_{w^s\sim p_{\theta}}[r(w^s)]$; $\nabla_{\theta}L_{RL} \approx -(r(w^s) - b)\nabla_{\theta}\log{p_{\theta{}}(w^s)}$, where $w^s$ is a description sampled from the model, $r(\cdot)$ is the reward function, and $b$ is the baseline.
We employ the SCST training strategy~\cite{rennie2017self} and use the reward for descriptions from the greedy decoding (i.e., $b=r(w^g)$) as the baseline. We also employ CIDEr as the reward, following \citet{rennie2017self}'s observation (using CIDEr as a reward improves overall metric scores). We follow the mixed loss strategy setup~\cite{wu2016google, paulus2018a}: $L = \gamma_1 L_{ML} + \gamma_2L_{RL}$.

\vspace{3pt}
\par
\noindent\textbf{Multilingual Parameter Sharing.}
We implement the multilingual training setup by sharing parameters between English and Hindi models, except the parameters of word embeddings, description LSTMs, and final fully connected layers, making the total number of parameters substantially less than those needed for the separate two models summed.  

\begin{figure}[t]
    \centering
    \includegraphics[width=0.95\columnwidth]{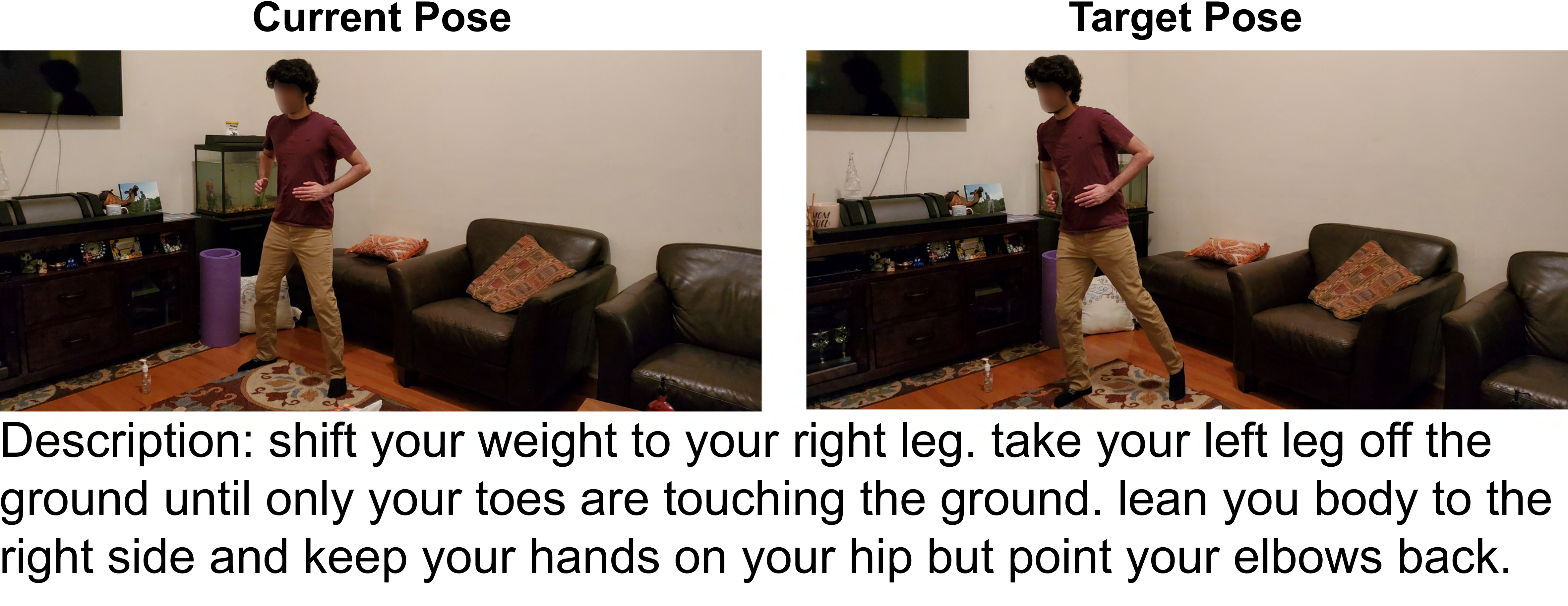}
    \caption{An example from Sim-To-Real transfer dataset.\label{fig:real_example}}
\end{figure}

\begin{table*}[t]
\begin{center}
\resizebox{1.85\columnwidth}{!}{
 \begin{tabular}{|c|l|c|c|c|c||c|c|c||c|c|}
  \hline
  \multirow{2}{*}{Language}  &  \multirow{2}{*}{Models} & \multicolumn{4}{c||}{Automated Metrics} & \multicolumn{3}{c||}{Task-Specific Metrics} & \multicolumn{2}{c|}{Human Eval.} \\
  \cline{3-11}
  && B4 & C & M & R& object-match & body-part-match & direction-match & R & F/G  \\
  \hline
  \hline
  
  \multirow{3}{*}{English} & V-Only & 6.90 & 6.41 & 16.78 &30.09& 0.04 & 1.01 & 0.05 & 4\% & 4\%\\
  & L-Only & 17.74 & 11.42 & 22.14 &35.16 & 0.08 & 1.22 & 0.15 & 15\% & 27\%\\
   
  & V+L & 17.55 & 14.47 & 21.29 & 35.21 & 0.18 & 1.29 & 0.13 & 48\% & 45\%\\
  \hline
  \multirow{3}{*}{Hindi} & V-Only & 8.43 & 4.37 & 18.90 & 28.55  &  0.03 & 1.21 & 0.02 & 9\% & 10\%\\
    & L-Only & 25.42 & 11.41 & 29.68 &36.90 & 0.0 & 1.42   & 0.07  &  19\% & 26\% \\
  & V+L & 18.99 & 8.58 & 29.26 & 34.73 & 0.08 & 1.63 & 0.10 & 51\%  & 53\%\\
  \hline

\end{tabular}
}
\end{center}
\caption{The performance of the unimodal and multimodal models on automated metrics, our new task-specific metrics, and human evaluation. for both English and Hindi dataset on the val-seen split (B4: BLEU-4, C: CIDEr, M: METEOR, R: ROUGE, V: Vision+Joints, L: Language, R: Relevancy, F/G: Fluency and Grammar). \label{tbl:forward_results} } 
\end{table*}

\subsection{Target Pose Retrieval Model}
The current and target candidate images are encoded the same way as the captioning model. A bidirectional \textrm{LSTM} encodes the descriptions: $c = \textrm{BiLSTM}(\hat{w})$.
Image features are aligned with description features via cross attention.
\begin{align}
    \tilde{c}^c, \tilde{f}^{t_i}, &\tilde{c}^{t_i}, \tilde{f}^{c} = \textrm{CA-Stack}(c, f^c, c, f^{t_i}) \label{eq1}\\
    k_{1i} &= \textrm{Self-Gate}([ \tilde{c}^c;\tilde{c}^{t_i}; \tilde{c}^c \odot \tilde{c}^{t_i}])\\
    g_{1i} &=\textrm{Self-Gate}([  \tilde{f}^{t_i};\tilde{f}^{c};  \tilde{f}^{t_i} \odot \tilde{f}^{c}]) \label{eq2}
\end{align}  
where \(\odot\) is the element-wise product (see Appendix~\ref{app:model} for details of the \(\textrm{Self-Gate}\)). For joints feature, we calculate the difference between the two joints set: $J^{dt_i} = W_j^{\top}(J^{t_i}-J^c)$; $J^{dc_i} = W_j^{\top}(J^c-J^{t_i})$.
We apply the same process that the image features go through (i.e., Eq. \ref{eq1}-\ref{eq2}) to get $k_{2i}$ and $g_{2i}$.
\begin{align}
    p_i &= W_p^{\top}[  k_{1i};g_{1i};  k_{1i} \odot g_{1i}]\\
    q_i &= W_q^{\top}[  k_{2i};g_{2i};  k_{2i} \odot g_{2i}]\\
    s_i &= W_s^{\top}[  p_i;q_i;  p_i \odot q_i]
\end{align}
The score \(s_i\) is calculated for each target candidate and the one with the highest score is considered as the predicted one: $\hat{t} = \textrm{argmax}([ s_0;s_1;...;s_9])$.

\begin{table}[t]
\begin{center}
\resizebox{0.90\columnwidth}{!}{
 \begin{tabular}{|c|l|c|c|c|c|}
  \hline
  Language  &  Models & B4 & C & M & R\\
  \hline
  \hline
  
  \multirow{4}{*}{English} & V+L & 17.55 & 14.47 & 21.29 & 35.21\\
  & (-) Joints & 17.39 & 13.79 & 21.35 & 34.86 \\
   
  & (+) RL & 18.69 & 16.04 & 22.35 &36.18\\
  & (+) Multi-L & 19.08  & 15.71 & 22.47 & 36.46\\
  \hline
  \multirow{4}{*}{Hindi} & V+L & 18.99 & 8.58 & 29.26  & 34.73  \\
    & (-) Joints & 18.23 & 7.93 & 27.55 & 34.12\\
  & (+) RL & 18.57 & 9.63 & 28.83 & 34.76\\
  & (+) Multi-L & 18.67   &  9.77  & 29.05 & 34.74 \\
  \hline
  
\end{tabular}
}
\end{center}
\caption{Model ablations on val-seen split (RL: reinforcement learning, Multi-L: multilingual).\label{tbl:forward_ablation}} 
\end{table}

\begin{table}[t]
\begin{center}
\resizebox{0.98\columnwidth}{!}{
 \begin{tabular}{|c|l|c|c|c|c|}
  \hline
  Dataset & Model & B4 & C & M & R \\
  \hline
  \hline
  \multirow{2}{*}{\shortstack[c]{Image Editing Request\\\citet{tan2019expressing}}} &DRA & 6.72 & 26.36 & \textbf{12.80} & 37.25\\
    \cline{2-6}
  & Ours & \textbf{7.88} & \textbf{27.70} & 12.53 & \textbf{37.56}\\
  \hline
  \multirow{2}{*}{\shortstack[c]{NLVR2\\ \citet{suhr2019corpus}}} & DRA & 5.00 & \textbf{46.41} & 10.37 & \textbf{22.94}\\
    \cline{2-6}
  & Ours & \textbf{5.30} & 45.09 & \textbf{10.53} & 22.79\\
  \hline
    \multirow{2}{*}{\shortstack[c]{CLEVR-Change (SC)\\ Park et al. (2019)}} & DUDA & 42.9 & 94.6 & 29.7 & -\\
    \cline{2-6}
  & Ours & \textbf{44.0} & \textbf{98.7} & \textbf{33.4} & 65.5\\
  \hline
\end{tabular}
}
\end{center}
\vspace{-5pt}
\caption{Our baseline V+L model performs competitively on other image-difference captioning datasets (DRA: Dynamic Relation Attention~\cite{tan2019expressing}, DUDA: Dual Dynamic
Attention Model~\cite{park2019robust}; SC = Scene Change). \label{tbl:otherdataset}}
 \vspace{-8pt}
\end{table}

\section{Experimental Setup}
\noindent\textbf{Data Splits.} For the \posecorrectiontask{} task, we split the dataset into train/val-seen/val-unseen/test-unseen following \citet{anderson2018room2room}. We assign separate rooms to val-unseen and test-unseen splits for evaluating model's ability to generalize to unseen environments. The number of task instances for each split is 5,973/562/563/593 (train/val-seen/val-unseen/test-unseen) and the number of descriptions is 5,973/1,686/1,689/1,779. For the \poseretrievaltask{} task, we split the dataset into train/val-unseen/test-unseen. In this task, ``unseen" means ``unseen animations". We split the dataset by animations so that the task cannot be easily done by memorizing/capturing patterns of certain animations in the image pairs. After filtering for the target candidates (see Sec. \ref{sec:pose_retri}), we obtain 4,227/1,184/1,369 (train/val-unseen/test-unseen) instances. See Appendix~\ref{app:data_splits} for the detailed room and animation assignments.
\vspace{3pt}
\par
\noindent\textbf{Training Details.}
We use 512 as the hidden size and 256 as the word embedding dimension. We use Adam \cite{kingma2014adam} as the optimizer. See Appendix~\ref{app:training} for details.

\vspace{3pt}
\par
\noindent\textbf{Metrics.}
For the \posecorrectiontask{} task, we employ automatic evaluation metrics: BLEU-4~\cite{papineni2002bleu}, CIDEr~\cite{vedantam2015cider}, METEOR~\cite{banerjee2005meteor}, and ROUGE-L~\cite{lin2004rouge}. Also, motivated by previous efforts towards more reliable evaluation~\cite{wiseman2017challenges, serban2017hierarchical, niu2019automatically, zhang2019bertscore, sellam2020bleurt}, we introduce new task-specific metrics to capture the important factors. Object-match counts correspondences of environment objects, body-part-match counts common body parts mentioned, and direction-match counts the (body-part, direction) pair match between the model output and the ground-truth (see Appendix~\ref{app:direction_match} for more information on direction-match). In the \poseretrievaltask{} task, we use the accuracy of the selection as the performance metric.
\vspace{3pt}
\par
\noindent\textbf{Human Evaluation Setup.} 
We conduct human evaluation for the \posecorrectiontask{} task models to compare the output of the vision-only model, the language-only model, and the full vision+language model qualitatively. We sample 100 descriptions from each model (val-seen split), then asked crowd-workers to vote for the most relevant description in terms of the image pair, and for the one best in fluency/grammar (or `tied'). Separately, to set the performance upper limit and to verify the effectiveness of our distractor choices for the \poseretrievaltask{} task, we conduct another human evaluation. We sample 50 instances from the \poseretrievaltask{} test-unseen split and ask an expert to perform the task for both English and Hindi samples. See Appendix~\ref{app:human_eval} for more details.

\vspace{3pt}
\par
\noindent\textbf{Unimodal Model Setup.}
We implement unimodal models (vision-/language-only) for comparison with the multimodal models. See Appendix~\ref{app:unimodal_model} for more details.

\vspace{3pt}
\par
\noindent\textbf{Other Image-Difference Datasets.}
We also evaluate our baseline model on other image-difference datasets to show that the baseline is strong and competitive: Image Editing Request~\cite{tan2019expressing}, NLVR2~\cite{suhr2019corpus} (the variant from~\citet{tan2019expressing}), and CLEVR-Change~\cite{park2019robust}.

\vspace{3pt}
\par
\noindent\textbf{Sim-to-Real Transfer.}
To verify the possibility of the transfer of our simulated image dataset to real images, we collect real image pairs of current and target poses. We randomly sample 60 instances from test-unseen split (test-sim) and then the authors and their family members\footnote{Hence covering diverse demographics, including some that are different from the simulator data splits, as well as different room environments. All participants consented to the collection of images (and additionally, we blur all faces).} follow the poses in the sampled test-sim split to create the real image version (test-real). Since the environments (thus objects and their layout too) and poses (though they are told to try to match as accurately as possible) have differences between the two splits (i.e., test-sim and test-real), we manually re-write a few words or phrases in the descriptions to make it more consistent with images in the test-real split (see Fig.~\ref{fig:real_example}).

\section{Results\label{sec:results}}

\begin{figure}[t!]
    \centering
    \includegraphics[width=0.95\columnwidth]{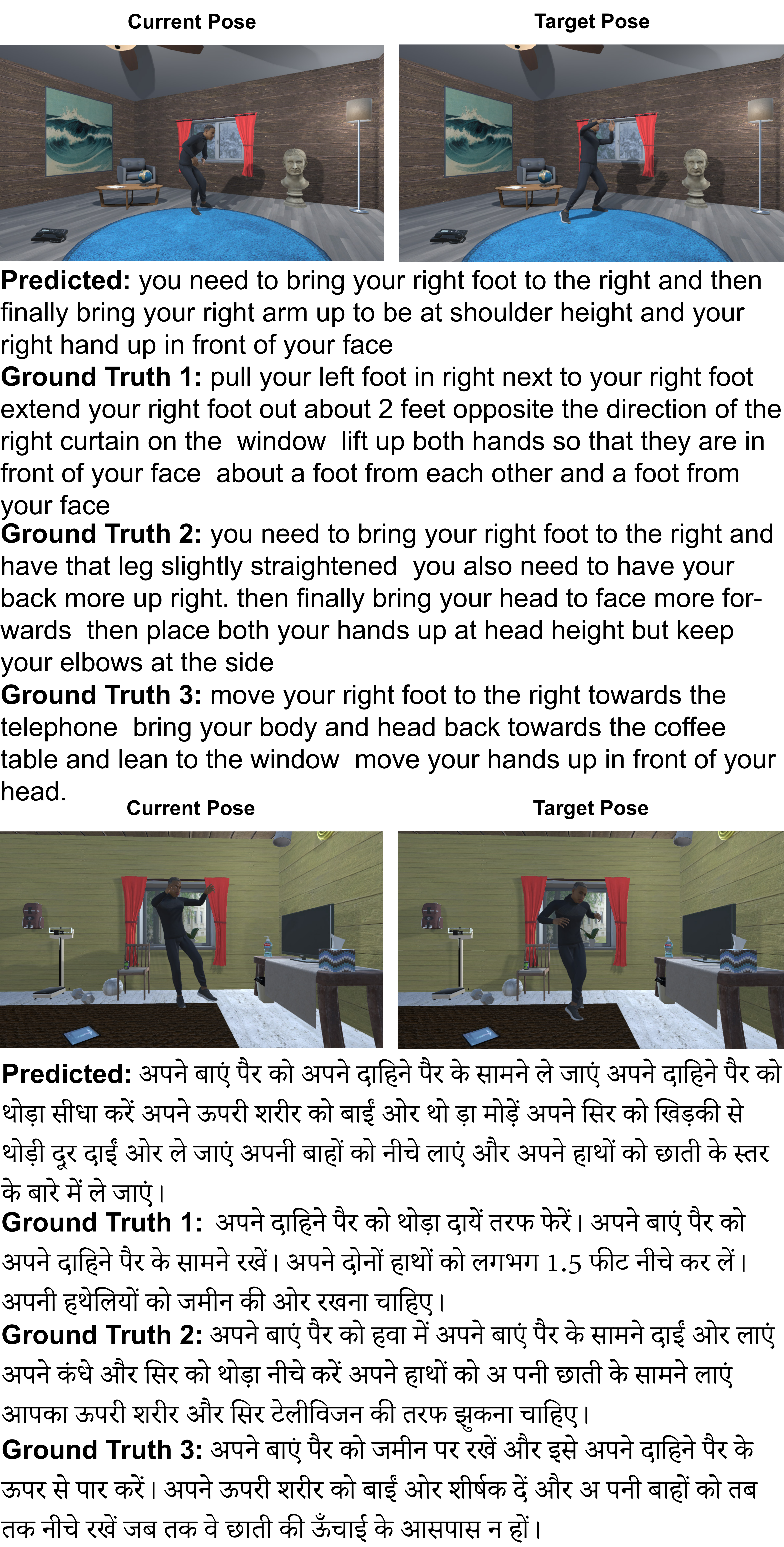}
    \caption{Output examples of our multimodal model in English (top) and Hindi (bottom).\label{fig:descriptionsfig}}
\end{figure}

\begin{table}[t]
\begin{center}
\resizebox{0.75\columnwidth}{!}{
 \begin{tabular}{|c|c|c|c|c|}
  \hline
  Character No. & B4 & C & M & R \\
  \hline
  \hline
  1  & 20.23 & 9.44 & 21.87 & 35.98 \\
   \hline
   2 & 17.54 & 7.43 & 20.20 & 34.70 \\
   \hline
   3 & 18.54 & 7.24 & 20.74 & 35.58 \\
    \hline
   4 & 19.00 & 9.28 & 20.43 & 34.01 \\
    \hline
   5 & 19.77 & 10.59 & 21.08 & 35.01 \\
   \hline
   6 & 20.28 & 7.94 & 20.94 & 35.47 \\
   \hline
  
\end{tabular}
}
\end{center}
\caption{The V+L model's performance (English) on the individual characters' demographics. The balanced scores indicate that our dataset is not biased towards any specific demographic.\label{tbl:demographic_bias}} 
\end{table}

\begin{table}[t]
\begin{center}
\resizebox{0.99\columnwidth}{!}{
 \begin{tabular}{|l|c|c|c|c||c|c|}
  \hline
   \multirow{2}{*}{Split} & \multicolumn{4}{c||}{Automated Metrics} & \multicolumn{2}{c|}{Task-Specific Metrics} \\
  \cline{2-7}
  & B4 & C & M & R & \;\;\;\;OM\;\;\;\; & DM \\
  \hline
  \hline
 
   test-sim & 16.93 & 9.91 & 21.79 & 35.08 & 0.04 & 0.20\\
   \hline
   test-real & 13.01 & 7.12 & 21.40 &33.05 & 0.07 & 0.11\\
   \hline
\end{tabular}
}
\end{center}
\caption{Sim-to-Real transfer performance. Since there is no GT joints for real images, the body-part-match metric is not available (OM: object-match, DM: direction-match). \label{tbl:real_image_result}} 
\end{table}

\subsection{Pose Correctional Captioning Task\label{sec:fwd_task_results}}
As shown in Table~\ref{tbl:forward_results}, the V+L models show better performance than V-only models. The L-only model shows higher scores on some of the automatic metrics, likely because the descriptions in our \dataname{} dataset are instructional about body parts (and their movements/directions), so similar phrases are repeated and shallow metrics will only focus on such phrase-matching, not correctly reflecting human evaluations~\cite{belz2006comparing, reiter2009investigation, scott2007nlg, novikova2017we, reiter2018structured}. Thus, we also evaluate the output of each model on our task-specific metrics that account the important factors (objects, body parts, and movement directions), and we also conduct human evaluation to check the real quality of the outputs. The V+L models show better performance on the task-specific metrics and human evaluation, meaning they capture essential information and their outputs are more relevant to the images and more fluent in the respective language. See Appendix~\ref{app:unseen_results} for ``unseen" split results.\footnote{We also checked for variance by running models with 3 different seeds and the stddev is small (less than/near 0.5\% on CIDEr).}

\vspace{3pt}
\par
\noindent\textbf{Ablations.} As Table~\ref{tbl:forward_ablation} shows, adding body joints features improves the score much, implying body joints gives additional important information to capture human movements.
\vspace{3pt}
\par
\noindent\textbf{RL/Multilingual Model Results.} As Table~\ref{tbl:forward_ablation} shows, RL training helps improve scores by directly using the evaluation metric (CIDEr) as the reward. We leave exploring more effective reward functions (e.g., the joints distance from a reverse pose generation task) for future work. Table~\ref{tbl:forward_ablation} also shows that the multilingual training setup achieves comparable scores (similar observation to \citet{wang2019vatex}) with only 71\% of the parameters of the separate training setup (13.2M vs 18.7M), promising future work on more compact and efficient multilingual models.

\vspace{3pt}
\par
\noindent\textbf{Other Image-Difference Datasets.}
Table~\ref{tbl:otherdataset} shows that our V+L baseline model beats or matches state-of-the-art models on other datasets, implying our baseline models are strong starting points for our \dataname{} dataset.

\vspace{3pt}
\par
\noindent\textbf{Output Examples.}
Outputs from our V+L models are presented in Fig.~\ref{fig:descriptionsfig}. The English model captures the movement of  the character's legs and arms (``bring your right foot to the right'' and ``bring your right arm up to be at shoulder height ... right hand up in front of your face''). The Hindi model captures movement of the body parts and their spatial relationship to each other (English translation: ``move your left leg in front of your right leg..."), the model can also describe movement using object referring expressions (English translation: ``...move your head slightly away from the window..."). See Fig.~\ref{fig:descriptionsfig} for the original Hindi and Appendix~\ref{app:output_examples} for full analysis and unimodal outputs.

\vspace{3pt}
\par
\noindent\textbf{Demographic Ablations.}
We split the dataset into subsets for each individual character avatar, and evaluate our V+L model on each subset. As shown in Table~\ref{tbl:demographic_bias}, scores from each subset are reasonably balanced, indicating our dataset is not skewed to favor a specific demographic or character.

\vspace{3pt}
\par
\noindent\textbf{Sim-to-Real Transfer.}
As shown in Table~\ref{tbl:real_image_result}, the sim-to-real performance drop is not large, meaning information learned from our simulated \dataname{} dataset can be transferred to real images reasonably well. Also, considering that the results are from a set of images of people with different demographics and different environments, there is no particular bias in the models' output which is trained on our dataset. Since there is no GT body joints for the real images, we modify our model so it can also be trained to predict the joints during training time as well as generate descriptions (i.e., in a multi-task setup) and use the estimated joints at test time.\footnote{For the simulated data results in Table \ref{tbl:forward_ablation} (English), we obtain a CIDEr score of 14.17 using predicted joints (on the val-seen split), which as expected is between the non-joint (13.79) and GT-joint (14.47) models’ results (hence showing that reasonable performance can be achieved without GT joint information at test time). The average distance between predicted and GT joints is around 0.4 meters.}

\begin{table}[t]
\begin{center}
\resizebox{0.62\columnwidth}{!}{
 \begin{tabular}{|l|c|c|}
  \hline
  \multirow{2}{*}{Models}  & \multicolumn{2}{c|}{Accuracy (\%)} \\
   \cline{2-3} 
   & English &Hindi \\
  \hline
  \hline
  Random-Selection &\multicolumn{2}{c|}{9.81} \\
   \hline
    V-Only &\multicolumn{2}{c|}{34.82} \\
   \hline
  L-Only & 8.86 & 8.96 \\
 \hline
  V+L & 38.49  & 37.84 \\
 \hline
 \hline
  Human & 96.00 & 96.00 \\
 \hline
\end{tabular}
}
\end{center}
\caption{The scores for the \poseretrievaltask{} task. While the V+L models scores the highest, there is still much room for improvement when compared with human performance. \label{tbl:retrieval_results} }
\vspace{-6pt}
\end{table}

\subsection{Target Pose Retrieval Task \label{sec:bwd_task}}
As shown in Table~\ref{tbl:retrieval_results}, V+L models show the highest scores for the \poseretrievaltask{} task, indicating that achieving high performance is not possible by exploiting unimodal biases. V-Only models score higher than the random-selection model, which selects an image at random, because even with our careful distractor choices (see Sec.~\ref{sec:pose_retri} and Appendix~\ref{app:distactor_choice}), the poses in the ``current'' and ``target" images are more similar to each other than the other images. However, the human-model performance gap is still quite large, implying there is much room for improvement.\footnote{Human performance is 96\% when given the full task (English), but much lower when only given lang. (38\%) or only vis. (22\%), further indicating that both lang.+vis. is needed to solve the task.}

\section{Conclusion and Future Work}
We introduced \dataname{}, a novel pose correctional description dataset in both English and Hindi. Next, we proposed two tasks on the dataset, \posecorrectiontask{} and \poseretrievaltask{}, both of which require models to understand diverse linguistic properties such as egocentric relation, environmental direction, implicit movement description, and analogous reference as well as capture fine visual movement presented in two images. We also presented unimodal and multimodal baselines as strong starter models. Finally, we demonstrated the possibility of transfer to real images. 
In future work, we plan to further expand the \dataname{} dataset with more languages and even more diversity in the character pool (e.g., height, age, etc. based on digital avatar availability) and animations. 

\section*{Ethics Statement}
Our paper and dataset hopes to enable people to improve their health and well-being, as well as strives to follow ethical standards, e.g., we especially try to maintain balance across diverse demographics and avoid privacy concerns by collecting data from a simulated environment (but still show good transfer to real images from authors), and we also expand beyond English so as to more inclusively cover multiple languages. Similar to other image captioning tasks/models, some imperfect descriptions from models trained on our FixMyPose dataset might also lead to difficult/unnatural poses. Presenting models’ confidence scores can help people ignore such unnatural pose corrections; however, most importantly, careful use is required for real-world applications (similar to all other image captioning models/tasks, e.g., the ones used for accessibility and visual assistance), and further broader discussion on developing fail-safe AI systems is needed.

\section*{Acknowledgments}
We thank the reviewers and Jason Baldridge, Peter Hase, Hao Tan, and other UNC-NLP members for their helpful comments. This work was supported by NSF Award 1840131, NSF-CAREER Award 1846185, DARPA MCS Grant N66001-19-2-4031, Microsoft Investigator Fellowship, and Google Focused Award. The views contained in this article are those of the authors and not of the funding agency.

\bibliography{aaai21.bib}

\appendix
\section*{Appendices}

\begin{figure*}[t]
    \centering
    \includegraphics[width=1.85\columnwidth]{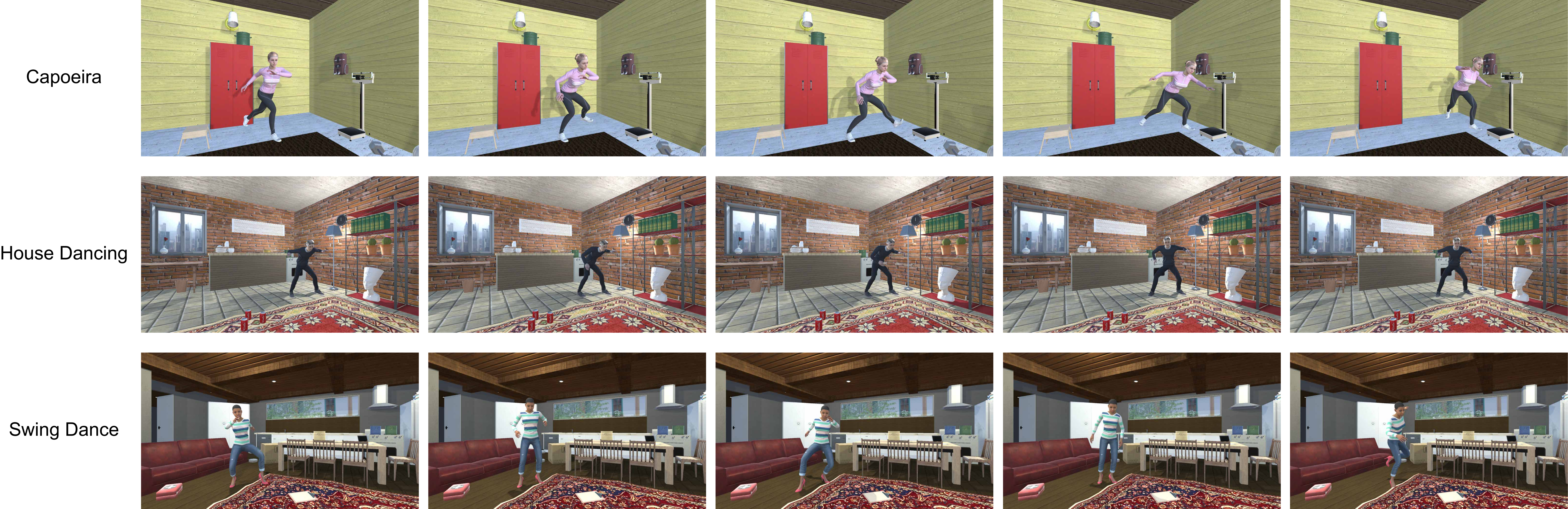}
    \caption{Examples of specific movement animations (each image is 10 frames apart). Each image sequence show a segment of the movement animation.}
    \label{fig:dances}
\end{figure*}

\begin{figure}[t]
    \centering
    \includegraphics[width=0.99\columnwidth]{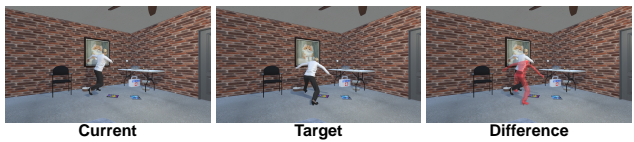}
    \caption{Current, target, and difference images. The target images are taken 10 frames after the current images are taken. The difference image shows the overlap of the ``current" and ``target" images with the pose in the ``target" image shown in red. \label{fig:before_after_diff}}
    
\end{figure}

\begin{figure*}[t]
    \centering
    \includegraphics[width=1.85\columnwidth]{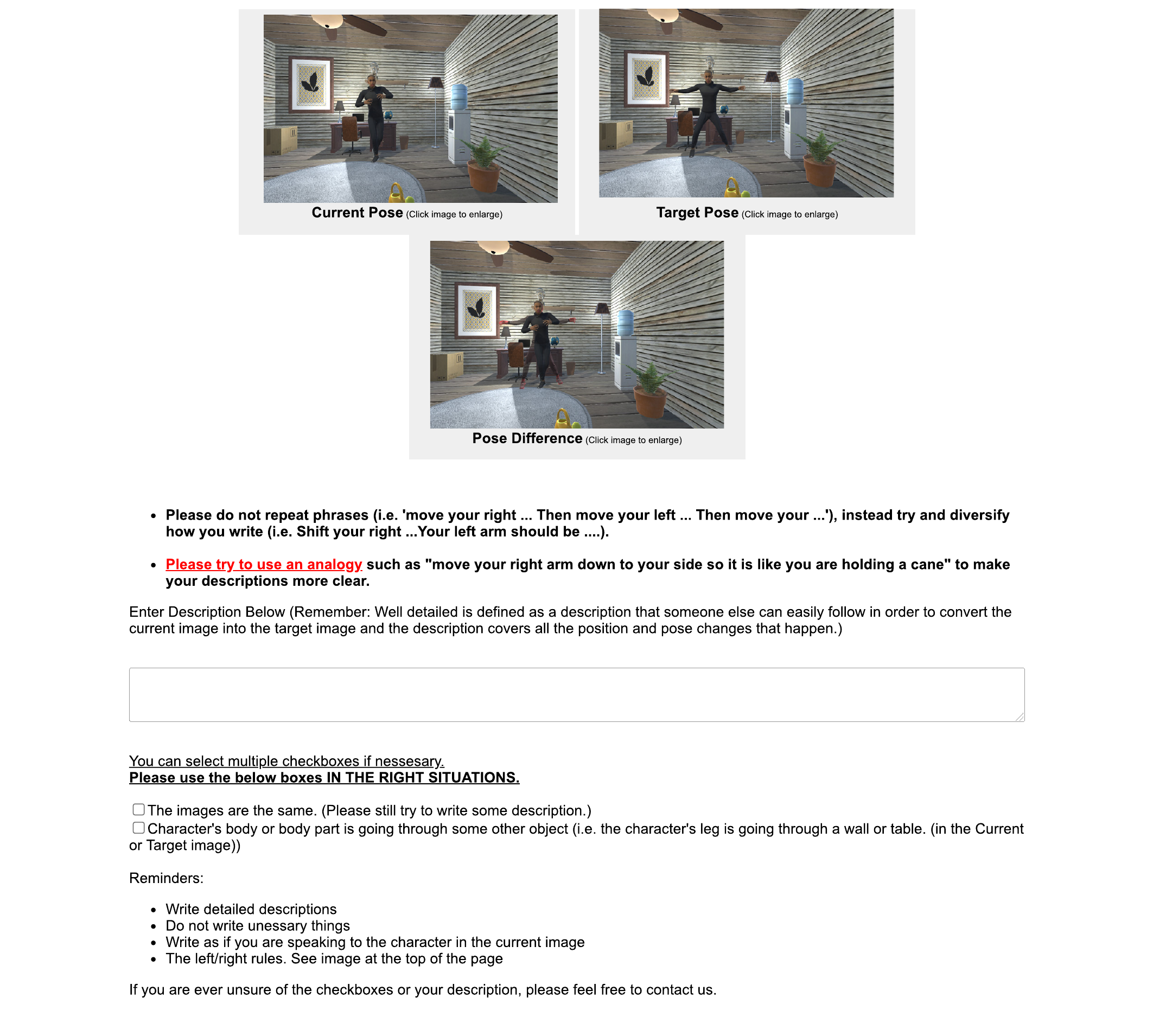}
    \caption{The interface of the writing task.}
    \label{fig:writing_interface}
\end{figure*}

\begin{figure*}[t]
    \centering
    \includegraphics[width=1.85\columnwidth]{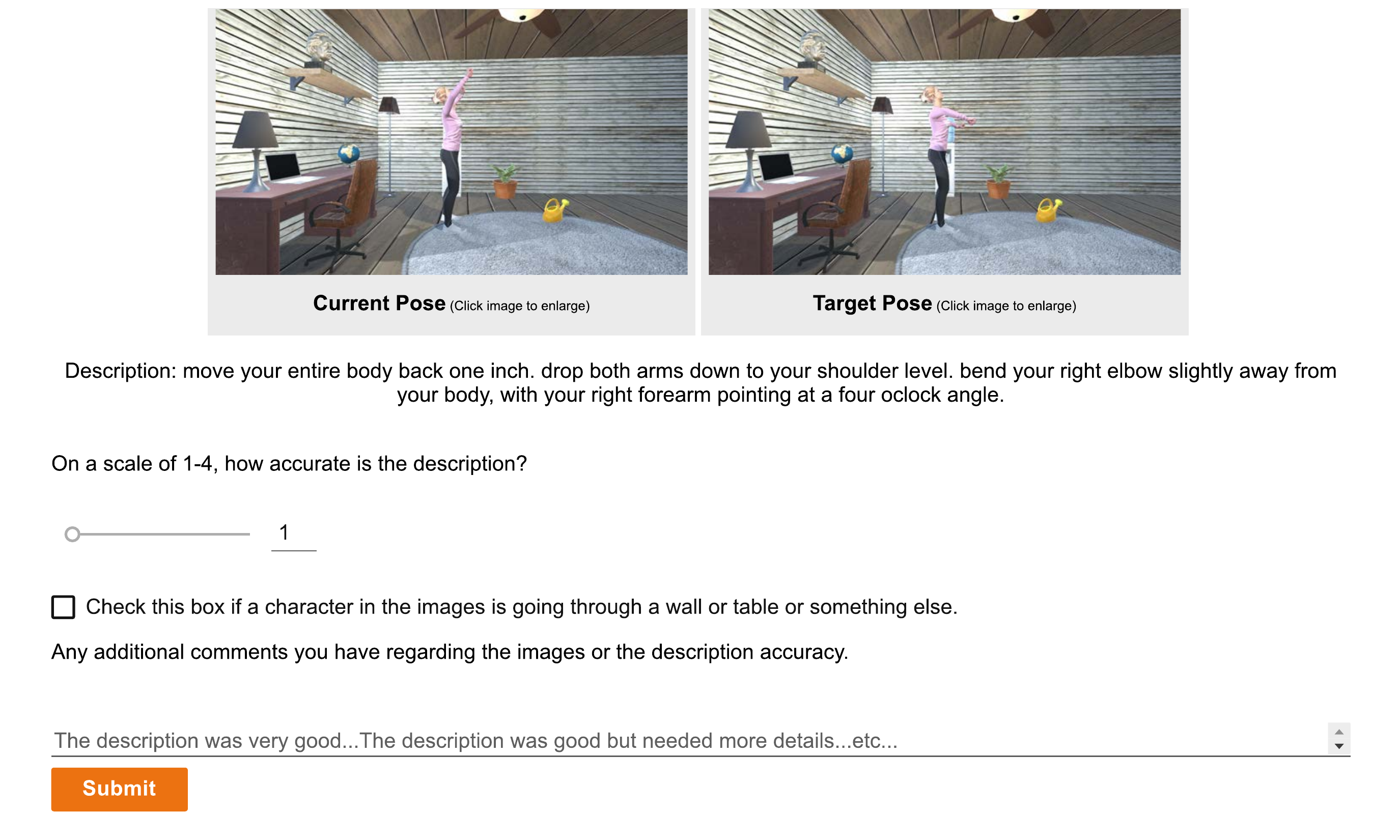}
    \caption{The interface of the verification task.}
    \label{fig:verification_interface}
\end{figure*}
\begin{figure*}[t]
    \centering
    \includegraphics[width=1.85\columnwidth]{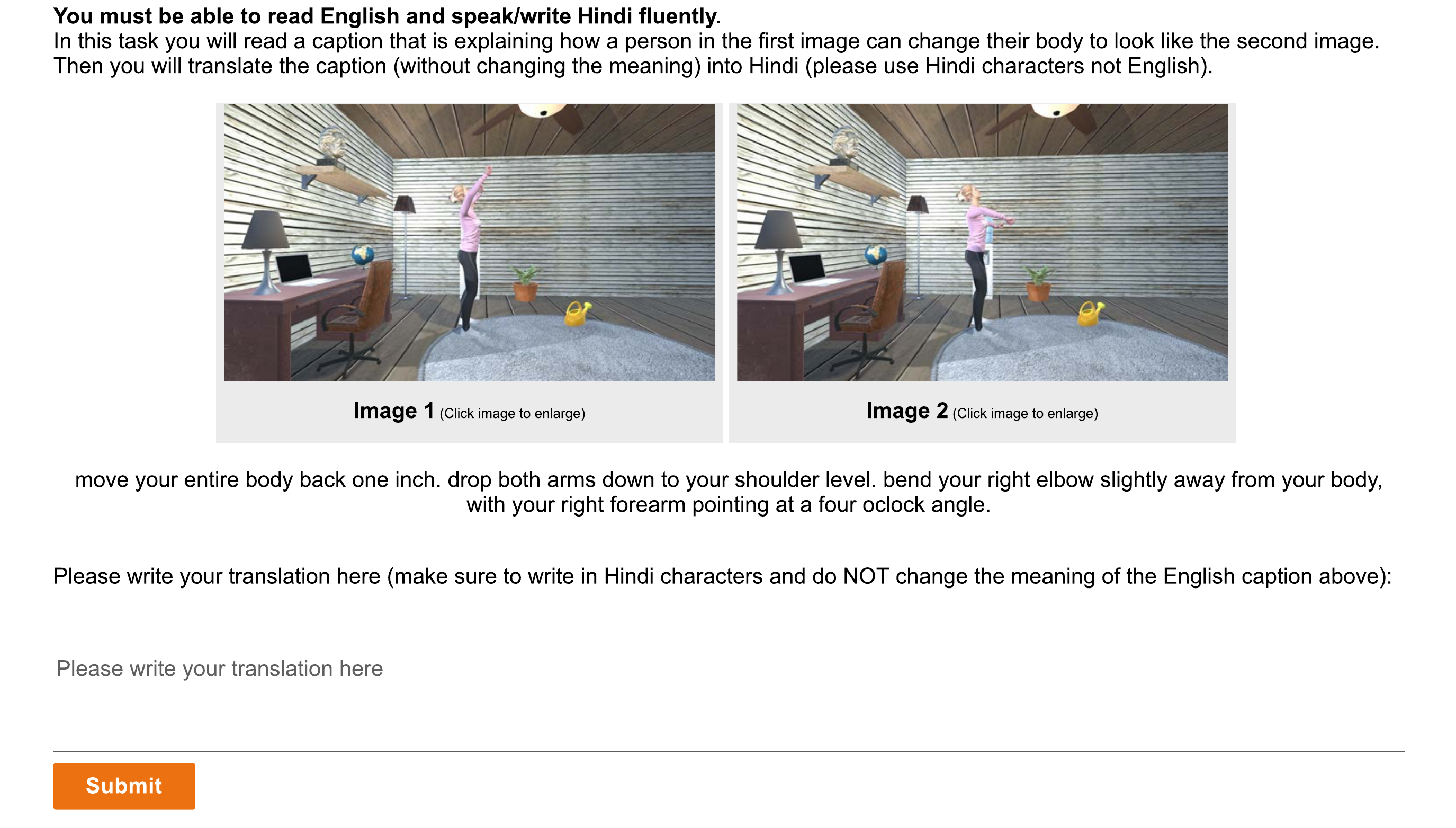}
    \caption{The interface of the translation task.}
    \label{fig:translation_interface}
\end{figure*}
\section{Distractor Choice Criteria\label{app:distactor_choice}}
For the ``target" and distractor images of the \poseretrievaltask{} task, we only consider images that meet these criteria: (1) the ``target" pose image must have more than 10cm average joints distance from the ``current" pose image, (2) each distractor has an average joints distance between 10cm and 1m from the ``target" pose image, (3) each distractor must have less than 2m average joints distance from the ``current" pose image, (4) the ``current" pose image is included in the distractor images, and (5) all the distractor images are from the same environment and have the same character as the one in the ``target" pose image.

\section{Dataset\label{app:dataset}}
\subsection{Image and Environment Generation \label{app:collection_details}}
\paragraph{Environment Creation.}
Every object inside of each room is collected from free assets in the Unity Asset Store\footnote{https://assetstore.unity.com} and various other free online resources. The first room is also collected as a free asset from the Unity Asset Store, however the rest of the rooms are manually created. In all rooms, including the first room, we manually choose and configure the arrangement of the objects.

\paragraph{Movement Animations.}
Movement animations are taken from realistic and natural body movements that people could potentially perform at home. Fig.~\ref{fig:dances} shows a few key frames of some movement animations. All characters and animations are collected from the free collection on Adobe's Mixamo.

\paragraph{Image Capture.}
To obtain each pair of images, we run the same animation twice but the second instance of the animation is offset by 10 animation frames. The 10 frame offset helps ensure that a clear visual difference is created, but not so much that it creates two unrelated images. The image of the first instance is the ``current" image and the image of the second instance is the ``target" image. Fig.~\ref{fig:before_after_diff} shows an example of a ``current" and ``target" image as well as a ``difference" image which shows the overlap of the ``current" and ``target" images with the pose in the ``target" image shown in red. Every 20 frames, an image pair is captured. 

\begin{figure}[t]
    \centering
    \includegraphics[width=0.35\columnwidth]{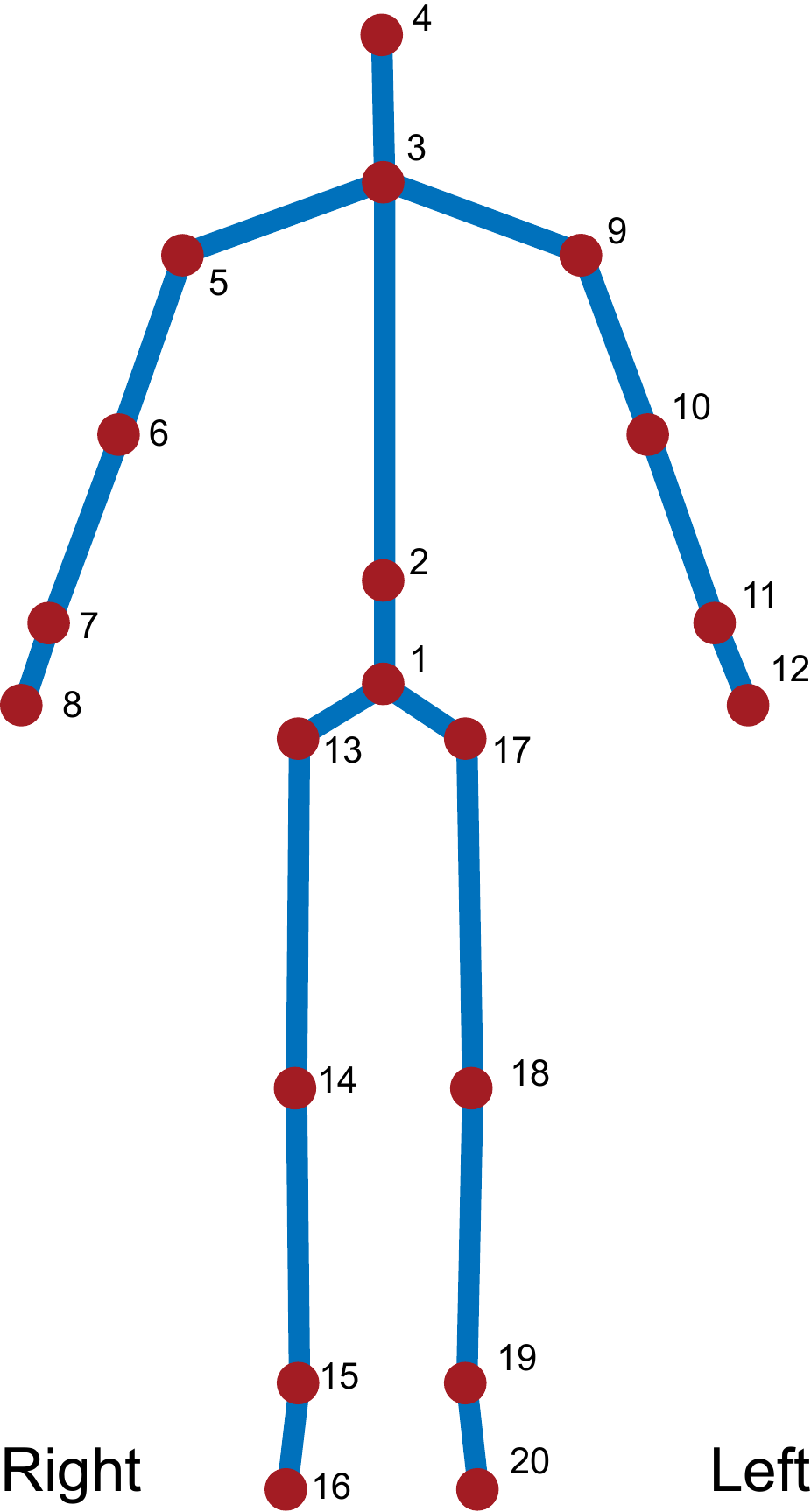}
    \caption{The 3D joint configuration of characters (from index 1 to 20: center hip, spine, neck, head, right shoulder/elbow/wrist/hand, left shoulder/elbow/wrist/hand, right hip/knee/ankle/foot, left hip/knee/ankle/foot).}
    \label{fig:joint_diagram}
\end{figure}

\paragraph{3D Body Joint Data.}
We obtain the 3D positional joint data of the character's poses from both the ``current" and ``target" images (see Fig.~\ref{fig:joint_diagram}). The positional data is relative to the camera's position and angle. This keeps all the data normalized regardless of which room or location in a room is chosen.

\begin{figure*}[t]
\centering
    \includegraphics[width=1.50\columnwidth]{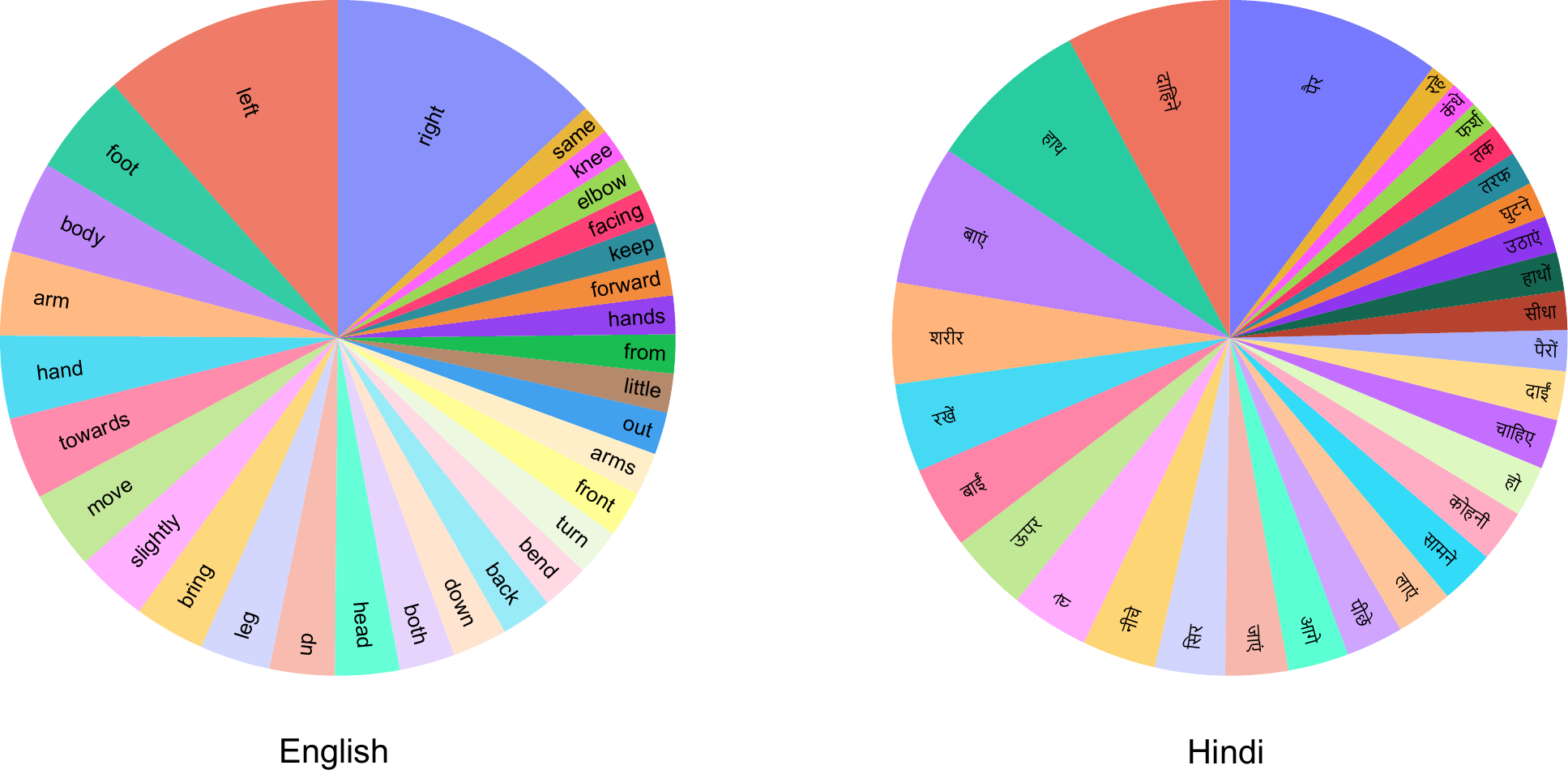}
    \caption{The 30 most common English/Hindi words in the dataset (excluding stop words). They primarily relate to directions, body parts, and movements.\label{fig:most_common_words}}
\end{figure*}
\begin{figure}[t]
    \centering
    \includegraphics[width=.99\columnwidth]{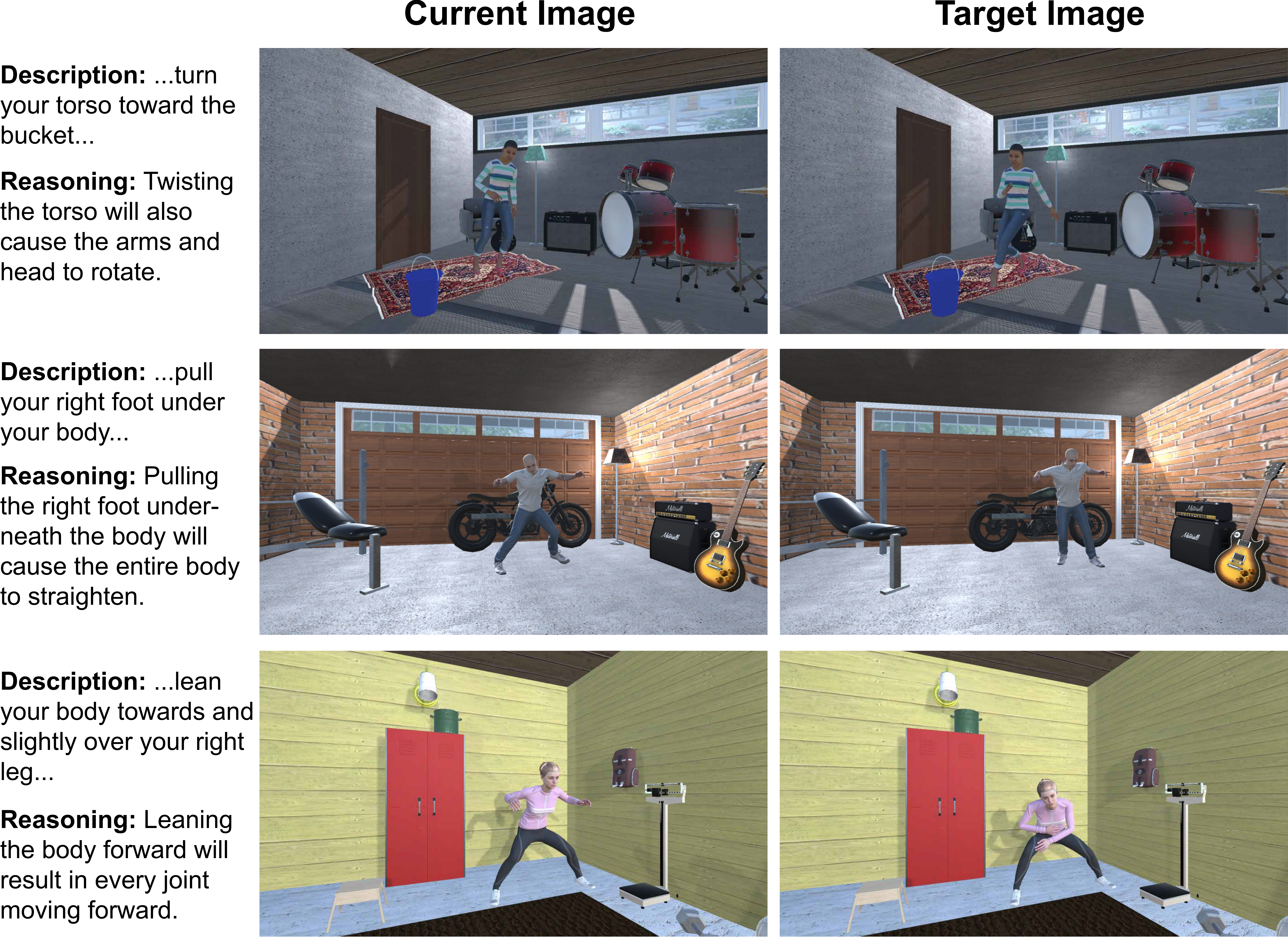}
    \caption{Examples of the `implicit movement description' linguistic property.}
    \label{fig:implied_move}
\end{figure}

\subsection{Data Collection Interface \label{app:interfaces}}
For each of the 3 data collection tasks (writing, verification, translation), we create a separate interface. The writing task and verification task are also provided with certain flags (detailed in corresponding interface paragraphs). Upon clicking the images in any of the interfaces, the clicked image will be enlarged and an option to view the image in a separate tab is given in case the worker would like an even larger image. 

\paragraph{Writing Task Interface.}
During this task, the goal is to have the workers see the 3 images (``current", ``target", ``difference") and then write a correctional description based on those images. The interface (as shown in Fig.~\ref{fig:writing_interface}) provides the 3 images labeled and a writing area. Workers are also provided with ``no clear difference" and ``character is going through an object" flag. The ``no clear difference" flag is designed to be used in the case the difference between the poses in the ``current" image and ``target" image is too small to write a good description. The ``character is going through an object" flag is meant to be used in the event that a character in either image has a body part going through a wall, table, or any other object.

\paragraph{Verification Task Interface. \label{app:verification_interface}}
This task serves to filter out any descriptions that are of poor quality. To do this, workers are provided with the ``current" image and the ``target" image and then the correctional description that is written for that image pair. They are then asked to rank the quality of the description from 1-4, with 1 being the description is completely unrelated and 4 being the description is perfect. Then, just as for the writing task, a checkbox for the ``character is going through an object" flag is provided in case the writing task workers miss it. The interface is shown in Fig.~\ref{fig:verification_interface}.

\paragraph{Translation Task Interface. \label{app:translation_interface}}
During this task, workers are asked to translate descriptions from English into Hindi. As shown in Fig.~\ref{fig:translation_interface}, the interface provides the ``current" and ``target" images for context and then the English description. Then, a text field is provided where workers can write the translation.

\subsection{Data Collection Filters \label{app:data_filters}}
During the writing task, some active quality checks are put in place to ensure that descriptions are of a certain base quality before they reach the verification task. Below is the list of each active quality that is put in place.
\begin{itemize}[leftmargin=*,noitemsep]
    \item Each description must contain at least 30 words.
    \item The symbols (, [, ], ), \&, *, \^{}, \%. \$, \#, and @ cannot be included.
    \item At least 50\% of the words in the instruction must be unique.
    \item The term ``image" cannot be included.
    \item The term ``i" cannot be included.
    \item The term ``target" cannot be included.
    \item The term ``difference" cannot be included.
\end{itemize}
In the case that workers in writing task select the ``no clear difference" checkbox on the interface, the 30-word minimum check is removed so that workers could write shorter descriptions, since there is not much difference to write about if the images are almost the same.

\subsection{Worker Qualifications and Incentives\label{app:worker_qual_payment}}
There were a total of 356, 373, and 47 unique crowd-workers who successfully passed the qualifications and completed the writing, verification, and translation tasks, respectively, at least once.\footnote{We do not collect or use any private information from the workers.}

\paragraph{Worker Qualifications.}
For all 3 tasks, crowd-workers are required to pass certain qualifications before they could begin. As both writing and verification tasks require reading (and writing in the case of the writing task) English, we require workers to be from native-speaking English countries and as the translation task requires translating to Hindi, we require that workers be from India. Crowd-workers are also required to have at least 1000 approvals from other tasks and a 95\% or higher approval rating.

\paragraph{Worker Payment. \label{app:worker_incentive}}
The writing task takes around 1 minute and workers are paid \$0.18 per description. For the first 25 high-quality descriptions that a worker writes, an additional bonus of \$0.02 is given for each description and then for every subsequent 50 high-quality descriptions written, the bonus per description is increased by \$0.01 (\$0.02 bonus per description for first 25, \$0.03 bonus for the next 50, \$0.04 bonus for the next 50, and so on). With this bonus rate, workers could get more than \$0.20 quite easily since the task is not long (and hence overall reasonably higher than minimum hourly wages). Since there is no limit on how much a worker can write, they could potentially keep stacking the bonus as much as they want.

\begin{table*}[t]
\begin{center}
\resizebox{1.95\columnwidth}{!}{
 \begin{tabular}{|c|c|c|c|}
 \hline
    Reference Frame & Freq. & Examples (English)\\
 \hline
   \multirow{3}{*}{Egocentric Relation} & \multirow{3}{*}{100\%} & ``... \textbf{rotate your left shoulder} so that your hand is \textbf{above your elbow} ..." \\
    &  & ``... put \textbf{your right foot and leg} forward so it is parallel \textbf{with your torso} ..." \\
    & & ``... move \textbf{your left leg} down and put in front of \textbf{your right leg} ..." \\
    \hline
   \multirow{3}{*}{Environmental Direction} &  \multirow{3}{*}{52\%} & ``... turn your left leg and right leg to the left to \textbf{face the wall with the door} ..." \\
   && ``...turn your head to \textbf{look to the bed}..." \\
   && ``...somewhat \textbf{aligning your eyes with the closest lamp} ..." \\
   \hline
   \multirow{3}{*}{Implicit Movement Description} &  \multirow{3}{*}{58\%} & ``... lean your body towards and slightly over your right leg ..." \\
    && ``... rotate your torso slightly to the left ..." \\
    && ``... then slightly lean forward ..." \\
    \hline
    \multirow{3}{*}{Analogous Reference} &  \multirow{3}{*}{18\%} & ``... extend your right arm straight in front of you \textbf{as if you are gesturing for someone to stop} ..." \\
    && ``... twist your upper body back to your right \textbf{in a golf swing motion} ..." \\
    && ``... hold your right hand next to your body \textbf{as if you are leaning on a cane} ..." \\
   \hline
\end{tabular}
}
\end{center}
\caption{Frequencies and detailed examples of the different properties present in correctional descriptions. \label{tbl:property_frequnecies_append}
} 
\end{table*}

\section{Analysis}
\subsection{Most Commonly Occurring Words\label{app:most_common_pie}}
The most commonly occurring words in our dataset are about direction, body parts, and movement, showing that models need to have a sense of direction with respect to body parts and objects, and also capture the differences between the poses to infer the proper movements. Fig.~\ref{fig:most_common_words} shows the most commonly occurring English/Hindi words in our dataset, which also primarily relate to directions, body parts, and movements.

\subsection{Description Length\label{app:description_length}}
The average length of the multi-sentenced descriptions (49.25 English / 52.74 Hindi words) is quite high, indicating that they are well detailed. The stddev (17.28/18.88) and the gap between the min and max (20/14 vs. 188/239) is quite large, reflecting the varying degrees of difference between the poses in an image pair. This length characteristic of the \dataname{} dataset requires models to generate descriptions without being redundant or insufficient in detail.

\subsection{Linguistic Properties\label{app:linguistic_properties}}
The descriptions in the \dataname{} dataset contain diverse linguistic properties. These properties as well as a few additional examples are provided in Table~\ref{tbl:property_frequnecies_append}. Additional examples of implicit movement description along with basic explanations are shown in Fig.~\ref{fig:implied_move}.

\section{Models \label{app:model}}
\paragraph{Cross Attention Stack.}
\(\textrm{CA-Stack}\) is a stack of cross attentions.

\begin{equation}
 \textrm{CA-Stack}(f^c, \hat{J}^c, f^t, \hat{J}^t):\left\{
\begin{aligned}
\bar{f}^c, \bar{J}^c &= \textrm{CA}(f^c, \hat{J}^c)\\
\bar{f}^t, \bar{J}^t &= \textrm{CA}(f^t, \hat{J}^t) \\
\tilde{f}^c, \tilde{J}^t &= \textrm{CA}(\bar{f}^c, \bar{J}^t)\\
\tilde{f}^t, \tilde{J}^c &= \textrm{CA}(\bar{f}^t, \bar{J}^c) \\
\end{aligned} 
\right. 
\end{equation}

where \textrm{CA} is cross attention.
\paragraph{Cross Attention.}
We calculate the similarity matrix, \(S\), between two features.
\begin{equation}
    S_{ij} = f^{\top}_ig_j
\end{equation}
From the similarity matrix, the new fused instruction feature is:
\begin{align}
    \hat{f} &= \textrm{softmax}(S^{\top})\cdot f \\
    \bar{g} &= W_g^{\top}[g;\hat{f};g \odot \hat{f}]
\end{align}
Similarly, the new fused visual feature is:
\begin{align}
   \hat{g} &= \textrm{softmax}(S)\cdot g \\
    \bar{f} &= W_{f}^{\top}[f;\hat{g};f \odot \hat{g}]
\end{align}
where \(W_g\) and \(W_f\) are trainable parameters, \(\odot \) is the element-wise product, and $\cdot$ is matrix multiplication. 

\paragraph{General Attention.}
We employ a basic attention mechanism for aligning description features, \(h\), and each of the visual and joints features.
\begin{align}
    A_{i} &= f_i^{\top}h\\
    \alpha &= \textrm{softmax}(A) \\
    \hat{f} &= \alpha^{\top}f 
\end{align}

\paragraph{Self Gate.}
We employ a basic attention mechanism for weighted summation of features.
\begin{align}
    A_{i} &= \textrm{Linear}(k_i)\\
    \alpha &= \textrm{softmax}(A) \\
    \hat{k} &= \alpha^{\top}k 
\end{align}

\section{Experiments}
\subsection{Data Splits\label{app:data_splits}} For the \posecorrectiontask{} task, we split the dataset into train/val-seen/val-unseen/test-unseen. Since each room in our \dataname{} has different visual setting (i.e., wall, floor, furniture, etc.), we assign separate rooms to val-unseen and test-unseen split. To be specific, we assign room 1 to 19, 24, and 25 to the train and val-seen splits, room 20 and 21 to the val-unseen, and room 22 and 23 to the test-unseen split. The final number of task instances for each split is 5,973/562/563/593 (train/val-seen/val-unseen/test-unseen) and the number of descriptions is 5,973/1,686/1,689/1,779. For the \poseretrievaltask{} task, we split the dataset into train/val-unseen/test-unseen. However, ``unseen" in this task means ``unseen animations". The reason we split the dataset by animations is that, otherwise, the task would be easier by memorizing/capturing some patterns in the image pairs from certain animations. We assign animation 6 and 16 to val-unseen, 7 to test-unseen, and the rest of the animations to the train split. After filtering for the target candidates, we obtain 4,227/1,184/1,369 (train/val-unseen/test-unseen) instances.

\subsection{Human Evaluation Setup\label{app:human_eval}}
We conduct human evaluation for the \posecorrectiontask{} task's models to compare the output of the V-only (V: vision+joints) model, the L-only (L: language) model, and the full V+L model qualitatively. We randomly sample 100 generated descriptions from each model (val-seen split), then asked 3 random crowd-workers (we also applied the standard quality filters of above 95\% hit success, over 1000 Hits, workers from native language-speaking countries) for each description to vote for the most relevant description in terms of the image pair, and for the one best in fluency/grammar (or `tied').\\
Separately, to set the performance upper limit and to verify the effectiveness of our distractor choices for the \poseretrievaltask{} task, we conduct human evaluation. We randomly sample 50 instances from the \poseretrievaltask{} test-unseen split and ask an expert for the English and Hindi samples to perform the task. Human evaluation is conducted the same way for both English and Hindi. We also ask the expert to complete the task from a unimodal perspective (i.e., only given the ``current" image or only given the description) to also show that the distractor choices cannot be exploited by any unimodal biases.

\begin{figure*}[t]
    \centering
    \includegraphics[width=1.99\columnwidth]{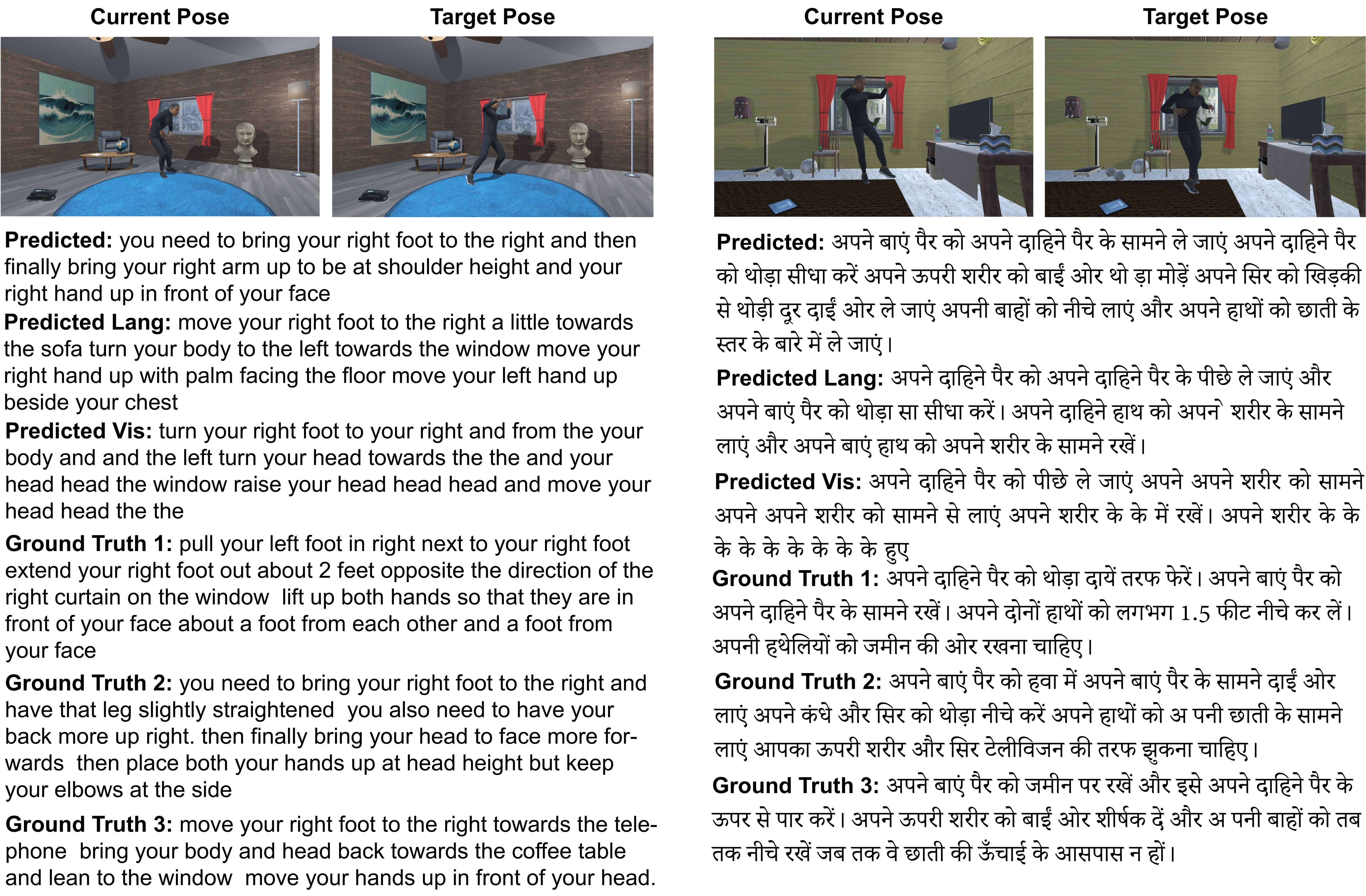}
    \caption{Output examples of our unimodal and multimodal models in English (left) and Hindi (right). ``Predicted" shows the V+L model output while ``Predicted Lang" and ``Predicted Vis" show the unimodal outputs for L-only and V-only models, respectively.\label{fig:outputexamples2}}
    
\end{figure*}

\subsection{Training Details (Reproducibility)\label{app:training}}
All of the experiments are run on a Ubuntu 16.04 system using the NVIDIA GeForce GTX 1080 Ti GPU and Intel Xeon CPU E5-2630. We employ PyTorch1.3 \cite{paszke2017automatic} to build our models (torchvision0.4/Python3.5/numpy1.18/scipy1.4).
The number of trainable parameters of the \posecorrectiontask{} V+L models are 9.1M and 9.6M for English and Hindi version, respectively (V-only: 10.4M/10.9M, L-only: 2.8M/3.1M), and the number of trainable parameters of the \poseretrievaltask{} V+L models are 10.9M/10.9M (V-only: 4.5M, L-only: 6.7M/6.7M). For the \posecorrectiontask{} task experiments, we use 9595/5555/2020 as the seed values, and run models 500 epochs and choose the best ones on val-seen/val-unseen splits. For the \poseretrievaltask{} task experiments, we use 5555/5556/5557 as the seed values, and run models 50 epochs and choose the best ones on the val-unseen split.
In the \posecorrectiontask{} task model training, at training time, the models are trained with teacher-forcing approach, and at test time, the greedy-search is employed to generate the descriptions. For the multilingual model, we freeze the shared parameters at the point at which the English score is the highest, and then fine-tune specific non-shared modules for each language with ML and RL training.
We employ ResNet-101 for the visual features. We use 512 as the hidden size and 256 as the word embedding dimension for both task models. We use the visual feature map of \(7\times7\) with 2048 channel size for the \posecorrectiontask{} task models and \(14\times14\) with 1024 channel size for the \poseretrievaltask{} task models. We use Adam \cite{kingma2014adam} as the optimizer and set the learning rate to $1\times 10^{-4}$ for ML training (for both tasks), and to $1\times  10^{-6}$ and $5\times 10^{-6}$ for RL training of English and Hindi models, respectively. The loss weights for ML+RL training ($\gamma_{1}$ and $\gamma_{2}$) are set to 0.05, and 1.0, respectively. For the dropout p value, 0.5 is used except for the multilingual training (0.3 is used). For hyper-parameters tuning, we try grid-search (e.g., dropout=\{0.3. 0.5\}, learning-rate=\{$1\times 10^{-4}$, ..., $1\times 10^{-6}$\}, etc).

\subsection{Direction-Match Metric\label{app:direction_match}}
We use the word order heuristic to extract (body-part, direction) pairs to compute direction-match. Our method can match 86\% and 87\% of human-extracted pairs for English and Hindi, respectively, meaning our metric is very closely matched with how humans would extract (body-part, direction) pairs.

\subsection{Unimodal Model Setup\label{app:unimodal_model}}
In the \posecorrectiontask{} task, the V-only model is not fed with the previous token at each decoding time step and does not attend to any previous tokens to decode the next token, and the L-only model does not take as input image pairs. In the \poseretrievaltask{} task, the V-only model selects the ``target" image only by comparing the ``current" image to distractors without the correctional description, the L-only model selects the ``target" image by comparing the correctional description to distractors without relying on the ``current" image.

\begin{table}[t]
\begin{center}
\resizebox{0.95\columnwidth}{!}{
 \begin{tabular}{|l|c|c|c|c||c|c|c|}
  \hline
   \multirow{2}{*}{Lang.} & \multicolumn{4}{c||}{Automated Metrics} & \multicolumn{3}{c|}{Task-Specific Metrics} \\
  \cline{2-8}
  & B4 & C & M & R& OM & BM & DM \\
  \hline
  \hline
  
 \multicolumn{8}{|c|}{Val-Unseen}\\
 \hline
   Eng.& 18.94 & 9.19 & 21.16 &35.04 & 0.11  & 1.59  & 0.18   \\
 \hline
   Hindi & 23.14  & 8.12  & 29.62  & 35.81  & 0.01 &  1.77   &  0.11   \\
  \hline
  \hline
   \multicolumn{8}{|c|}{Test-Unseen}\\
 \hline
  Eng. & 17.26 & 6.40 & 21.30 & 34.82 & 0.04 & 1.42 & 0.17   \\
  \hline
  
   Hindi & 18.98  & 6.69  & 28.47  & 34.53 & 0.03  & 1.52   & 0.11  \\

  \hline

\end{tabular}
}
\end{center}
\vspace{-6pt}
\caption{Val-unseen and Test-unseen: the performance of multimodal models on traditional automated metrics and our new task-specific metrics for both English and Hindi dataset (OM: object-match, BM: body-part-match, DM: direction-match). \label{tbl:forward_results_test_unseen}} 
\end{table}

\section{Results \label{app:othermetirc}}

\subsection{Output Examples\label{app:output_examples}}
Outputs from our V+L multimodal models are presented in Fig.~\ref{fig:outputexamples2}. Our multimodal English model captures the movement of  the character's legs and arms (``bring your right foot to the right'' and ``bring your right arm up to be at shoulder height ... right hand up in front of your face''). The Hindi model captures movement of the body parts and their spatial relationship to each other (English translation: ``move your left leg in front of your right leg..."), the model can also describe movement using object referring expressions (English translation: ``...move your head slightly away from the window..."). See Fig.~\ref{fig:outputexamples2} for the original Hindi. For all of the unimodal models, the outputs perform poorly and do not accurately match the image pair. For the V-only models' outputs, the grammar and sentence structure are also very poor.

\subsection{``Unseen" Split Results\label{app:unseen_results}}
Table \ref{tbl:forward_results_test_unseen} shows our V+L models' scores on the val-unseen and the test-unseen splits (the scores are chosen by the best performance on the val-unseen split). We suggest that model tuning/selection be done on the val-seen/unseen splits and the results from the test-unseen are reported, following the practice of \citet{anderson2018room2room}.

\end{document}